\pdfoutput=1
\documentclass[11pt, table, dvipsnames]{article}

\usepackage[preprint]{acl}

\usepackage{times}
\usepackage{latexsym}

\usepackage[T1]{fontenc}
\usepackage[utf8]{inputenc}

\usepackage{microtype}

\usepackage{inconsolata}

\usepackage{graphicx}
\usepackage{bbding}
\usepackage[table]{xcolor}
\usepackage{soul}
\usepackage{tikz}
\usetikzlibrary{shapes, positioning}
\usepackage{listings}
\usepackage{placeins}
\usepackage{enumitem}
\usepackage{fancyhdr}

\definecolor{lightgrayish}{rgb}{0.97, 0.97, 0.97}
\definecolor{error1}{RGB}{252, 229, 205}
\definecolor{error2}{RGB}{244, 204, 204}
\definecolor{error1_border}{RGB}{255, 200, 140}
\definecolor{error2_border}{RGB}{255, 155, 155}
\definecolor{increase_green}{RGB}{120, 203, 106}
\definecolor{span_correct}{RGB}{66, 150, 103}
\definecolor{span_incorrect}{RGB}{234, 91, 111}
\definecolor{explanation_partial}{RGB}{88, 160, 200}
\definecolor{explanation_correct}{RGB}{72, 120, 164}
\definecolor{explanation_incorrect}{RGB}{255, 173, 96}

\newcommand{\hlc}[2][yellow]{{%
    \colorlet{foo}{#1}%
    \sethlcolor{foo}\hl{#2}}%
}

\definecolor{applegreen}{rgb}{0.55, 0.71, 0.0}
\newcommand{\greencheck}{\textcolor{applegreen}{\CheckmarkBold}}
\newcommand{\redx}{\textcolor{red}{\XSolidBrush}}
\newcommand{\shade}{\cellcolor{gray!10}}

\usepackage{booktabs}
\usepackage{multirow}
\usepackage{array}
\usepackage{arydshln}
\usepackage[normalem]{ulem}

\newcommand{\thickuline}[1]{%
    \bgroup
    \markoverwith{\rule[-0.9ex]{2pt}{1.0pt}}%
    \ULon{#1\hfill\mbox{}}%
}

\title{\textsc{OpeNLGauge}: An Explainable Metric for NLG Evaluation with Open-Weights LLMs}

\author{Ivan Kartáč {\normalfont \and}  Mateusz Lango {\normalfont \and} Ondřej Dušek \\
Charles University, Faculty of Mathematics and Physics \\ Institute of Formal and Applied Linguistics \\ Prague, Czechia \\
\texttt{\{kartac,lango,odusek\}@ufal.mff.cuni.cz}
}

\fancypagestyle{firstpage}{
  \fancyhf{}
  \fancyhead[L]{\vspace{2mm} \thickuline{Published as a conference paper at INLG 2025} \\}
  \fancyfoot[C]{\thepage}

}

\begin{document}
\maketitle
\thispagestyle{firstpage}

\begin{abstract}
Large Language Models (LLMs) have demonstrated great potential as evaluators of NLG systems, allowing for high-quality, reference-free, and multi-aspect assessments.
However, existing LLM-based metrics suffer from two major drawbacks: reliance on proprietary models to generate training data or perform evaluations, and a lack of fine-grained, explanatory feedback. 
We introduce \textsc{OpeNLGauge}, a fully open-source, reference-free NLG evaluation metric that provides accurate explanations based on individual error spans. \textsc{OpeNLGauge} is available as a two-stage ensemble of larger open-weight LLMs, or as a small fine-tuned evaluation model, with confirmed generalizability to unseen tasks, domains and aspects.
Our extensive meta-evaluation shows that \textsc{OpeNLGauge} achieves competitive correlation with human judgments, outperforming state-of-the-art models on certain tasks while maintaining full reproducibility and providing explanations more than twice as accurate. 
\end{abstract}

\section{Introduction}
Evaluating Natural Language Generation (NLG) systems remains a challenging research problem. Traditional overlap-based metrics, such as BLEU~\cite{papineni2002bleu} and ROUGE~\cite{lin-2004-rouge}, are still widely used but exhibit limited correlation with human judgments, particularly when assessing modern NLG systems~\cite{novikova-etal-2017-need}. With the rise of pre-trained language models, the research community began to shift toward model-based metrics that better capture semantic similarity, yet their performance was still unsatisfactory~\cite{yan-etal-2023-bleurt,glushkova-etal-2023-bleu}.

\begin{figure}[ht]
    \centering
    \includegraphics[width=.8\linewidth, trim=15 18 15 31, clip]{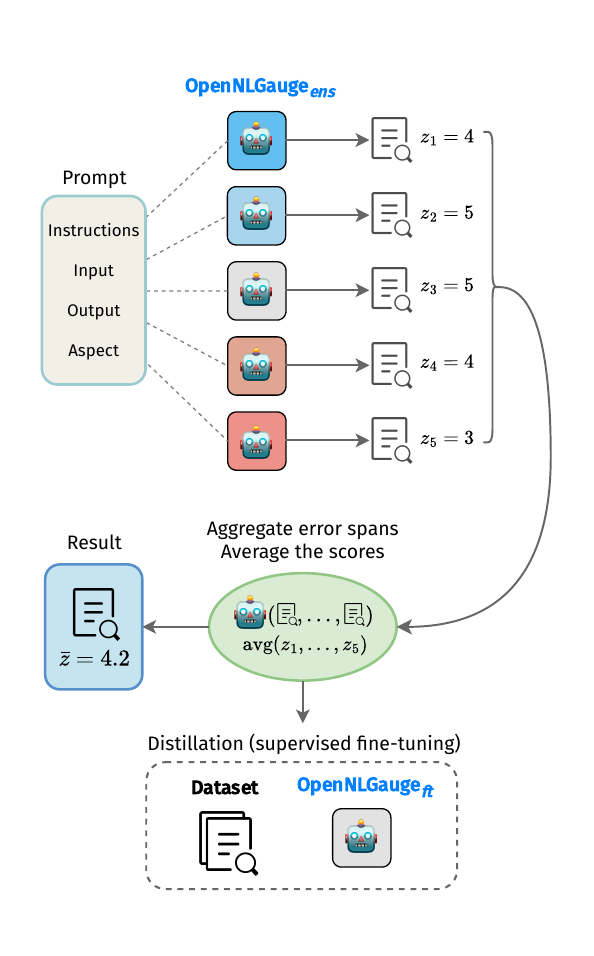}
    \caption{The ensemble metric \textsc{OpeNLGauge$_{ens}$} and its distilled version \textsc{OpeNLGauge$_{ft}$}.}
    \label{fig:ensemble_diagram}
\end{figure}

Recently, Large Language Models (LLMs) have demonstrated remarkable potential in imitating human evaluation of generated text~\cite{jiang2023tigerscore,xu-etal-2023-instructscore,hu2024themis}. 
LLM-based metrics are often general enough to evaluate diverse NLG tasks and can provide high evaluation performance without the need for reference texts~\cite{liu-etal-2023-g}. They also offer better adaptability to evaluate specific aspects of generated text and are able to evaluate beyond semantic correctness, taking into account aspects such as factual consistency or relevance~\cite{gu2025surveyllmasajudge}.
However, existing LLM-based NLG metrics lack two important features: (1) they are not fully open-source, with most relying on data generated by proprietary LLMs such as GPT-4 \cite{openai2024gpt4technicalreport}, and (2) they do not provide precise explanations for their evaluations, with most of them producing only overall scores, or explanations limited to a single short comment (see Table~\ref{tab:m-comparision} for details). 

In this paper, we introduce \textsc{OpeNLGauge} -- a~versatile, reference-free metric for NLG tasks that provides precise, error-span-based explanations (Figure~\ref{fig:ensemble_diagram}). Unlike existing LLM-based metrics, \textsc{OpeNLGauge} is built entirely on open-weight models and does not rely on human-annotated datasets. It supports a wide range of NLG tasks, including data-to-text, summarization or story generation. Moreover, it allows for fine-grained evaluation of specific, customizable aspects of the generated text such as faithfulness, fluency or coherence. It can also evaluate other user-defined aspects appropriate for a given application.

Our contributions are as follows:
\begin{itemize}[left=0pt,itemsep=2pt,parsep=2pt]
\item We introduce an effective prompting strategy and a two-stage LLM ensemble to obtain high-quality evaluations of NLG outputs.
The proposed \textsc{OpeNLGauge$_{ens}$} demonstrates higher correlations with human judgments on some NLG tasks than previously proposed metrics based on proprietary LLMs.
\item We collect outputs from over 100 NLG systems on 15 NLG tasks and use \textsc{OpeNLGauge$_{ens}$} to annotate them to construct a comprehensive dataset for training open NLG metrics.
\item The collected dataset is used to construct \textsc{OpeNLGauge$_{ft}$} -- a fine-tuned version of the smallest model from the Llama 3.1 family~\cite{grattafiori2024llama3herdmodels}, which is able to provide accurate explanations and reliable quality assessments of NLG outputs in a more cost-effective way.
\item We perform an extensive meta-evaluation of the proposed metrics on seven datasets covering five NLG tasks. The experimental analysis includes human evaluation of explanation quality, comparison of different score aggregation methods, ablation experiments, and evaluation on tasks, domains, and aspects not seen during training.
\end{itemize}
Our experiments show that \textsc{OpeNLGauge} achieves higher explanation quality than other metrics trained on data generated by proprietary LLMs, while also delivering strong evaluation performance. On tasks such as summarization, \textsc{OpeNLGauge} achieves higher correlations with human judgments than metrics based on the strong proprietary GPT-4 model. While the state-of-the-art metric, Themis -- which leverages both human-annotated data and data from proprietary LLMs -- remains superior in terms of correlations with human judgments, \textsc{OpeNLGauge} proves highly competitive and even outperforms it on certain tasks, while also providing more detailed error explanations.

All our experiments were conducted using only quantized open-weight models and two GPUs with 48GB of VRAM each, ensuring that the results can be reproduced in many AI research labs.
Our code and data are available on Github.\footnote{\url{https://github.com/ivankartac/OpeNLGauge}}

\section{Related Work}

\begin{table}[]
    \centering\small
    \renewcommand{\tabcolsep}{3.4pt}
    \begin{tabular}{lcccc}\toprule
    \bf Metric& \bf No Ref.& \bf Aspects & \bf Open & \bf Err. Span\\\midrule
    G-Eval& \greencheck  & \greencheck  & \redx& \redx\\
    Prometheus& \redx  & \greencheck  & \redx& \redx\\
    Auto-J& \greencheck  & \redx& \redx& \redx\\
    InstructScore& \redx  & \redx& \redx& \greencheck\\
    TIGERScore& \greencheck  & \redx& \redx& \greencheck\\
    Themis & \greencheck  &\greencheck  & \redx & \redx \\ \hdashline[0.5pt/2pt]
    \textsc{OpeNLGauge} & \greencheck  &\greencheck & \greencheck & \greencheck \\\bottomrule
    \end{tabular}
    \caption{Comparison of properties in different LLM-based metrics for NLG: reference-free (No Ref.), customizable aspects (Aspects), built exclusively using open-weight LLMs (Open), and producing precise explanations with error-span annotation (Err. Span). The metrics compared are: G-Eval~\cite{liu-etal-2023-g}, Prometheus~\citep{kim2023prometheus}, Auto-J~\citep{li2023generative}, InstructScore~\citep{xu-etal-2023-instructscore}, TIGERScore~\citep{jiang2023tigerscore}, Themis~\citep{hu2024themis} and \textsc{OpeNLGauge} (this work).
    } 
    \label{tab:m-comparision}
\end{table}

\begin{figure*}
    \centering\small
    \begin{tikzpicture}[align=left]

        \node[draw=gray!30, rectangle, line width=0.6mm, text width=15.5cm, anchor=north west, align=left, fill=gray!7.5, rounded corners] (context) at (-7.5, 4) {
            \textbf{Context:} \emph{[SUMMARY OF THE STORY]} \\[0.1cm]
            \textbf{Question:} What is Dr. Heywood Floyd's mission on the Clavius Base? \\[0.1cm]
            \textbf{Answer:} \hlc[error1]{According to the summary provided}\textsuperscript{[1]}, Dr. Heywood Floyd's mission on the Clavius Base is to investigate a~recently found artifact buried four million years ago. \hlc[error2]{Specifically, his mission is to ride in a Moonbus to the artifact, which is a monolith identical to the one encountered by the man-apes in the summary.}\textsuperscript{[2]}
        };

        \node[anchor=north west] (boxcontainer) at ([yshift=-0.3cm, xshift=0.1cm]context.south west) {
                    \textbf{Evaluation aspect:} \emph{Conciseness} -- Extent to which the answer is concise and to the point
        };

        \node[draw=error1_border, rectangle, line width=0.25mm, text width=7.3cm, below=0.6cm of boxcontainer.north west, anchor=north west, align=left, fill=error1, xshift=3pt] (box1) {
            \textbf{[1] Explanation:} This introductory phrase is an error affecting conciseness as it adds unnecessary words to the answer. The response could directly state Floyd's mission without referencing the source (summary), given the context of the question. \\
            \textbf{Severity:} 2
        };

        \node[draw=error2_border, rectangle, line width=0.25mm, text width=7.3cm, right=0.2cm of box1.north east, anchor=north west, align=left, fill=error2] (box2) {
            \textbf{[2] Explanation:} This phrase is considered an error affecting conciseness because it provides additional, unnecessary details about the mission. The initial sentence already clearly states Floyd's mission, making the subsequent elaboration redundant and wordy. \\
            \textbf{Severity:} 3
        };

        \node[rectangle, text width=15cm, below=0.1cm of box1.south west, anchor=north west, align=left] (score) {
            \textbf{Overall score:} Fair (3/5) \\[0.1cm]
            \textbf{Explanation of the score:} While the generated answer generally addresses the question, its conciseness is compromised by the inclusion of redundant information and an unnecessary introductory phrase. Removing these elements would enhance the answer's directness and efficiency, potentially elevating the score to `Good'.
        };

        \draw[thick, color=gray!60] ([xshift=-5pt, yshift=0.7cm] box1.north west) 
            rectangle 
            ([xshift=8pt, yshift=-5pt] score.south east);
        
    \end{tikzpicture}
    \caption{Example error span annotation provided by \textsc{OpeNLGauge} for the narrative question answering task. The answer to the question, grounded in the story summary, is evaluated for \emph{conciseness}.}
    \label{fig:exp1}
\end{figure*}

Although NLG has traditionally been evaluated using simple word-overlap-based metrics such as BLEU, these are known to have low correlations with human judgments \cite{novikova-etal-2017-need,reiter2018structured}.
This improved somewhat with the use of trained models for metrics in the past few years \cite{yuan2021bartscore,zhong2022towards,mehri2020usr}, but the correlations remained moderate.
Recently, numerous studies explored the application of LLMs in NLG evaluation.
A prominent line of research focuses on leveraging proprietary LLMs such as GPT-4 \cite{openai2024gpt4technicalreport}, with direct prompting for an overall score \cite{fu2024gptscore,kocmi2023large} or even annotating error spans with categories, which has been explored in machine translation \cite{kocmi2023gemba,fernandes2023devil,lu2023error}.
However, using proprietary models is costly and comes with a reproducibility penalty, as some LLM versions become unavailable or are modified in a non-transparent way~\citep{Chen2024How}. Another concern is data leakage, which affects results, but cannot be controlled in the case of proprietary models~\cite{balloccu2024leak}. 

Many recent LLM-based metrics, such as Themis~\citep{hu2024themis}, TIGERScore~\citep{jiang2023tigerscore}, InstructScore \cite{xu-etal-2023-instructscore} or Prometheus \cite{kim2023prometheus}, are built on open LLMs, but in fact they all rely on proprietary LLMs such as GPT-4~\cite{openai2024gpt4technicalreport} to generate, filter, or annotate their training data. Therefore, some of these metrics can be viewed as sophisticated knowledge distillation methods from proprietary to open-weight LLMs. 
This retains the reproducibility disadvantage, as reconstructing the metric from scratch or adapting it to a new task requires access to closed-source LLMs.

While several metrics based on open-weight LLMs provide some level of interpretability, this is often limited to a short, free-text review of the evaluated output \cite{hu2024themis,kim2023prometheus}.
Fine-grained error-span annotation offers several advantages over singular scores and comments:
(1) it can be easily processed automatically, allowing its use in post-processing steps or to provide feedback to the model or training algorithms, (2) it offers greater precision and clarity, as errors are associated with a particular part of the output, making it easier to find and correct issues, (3) it is more human-like, resembling the output of human annotation schemes such as MQM~\citep{10.1162/tacl_a_00437} or ESA~\citep{kocmi2024error}. The alignment with human evaluation allows easier comparison to humans or even use as pre-annotation, helping to accelerate human annotation process.
While some metrics do provide annotation on the error level, these tend to use closed LLMs and their scope is generally limited to single NLG tasks, such as the machine translation approaches mentioned above or \citet{kasner2024beyond}'s work on data-to-text generation.

Scoring outputs with a single LLM may introduce bias (\citealp{zheng2023judging, panickssery2024llm}), which can be alleviated by combining multiple LLMs as evaluators. \citet{verga2024replacingjudgesjuriesevaluating} apply an ensemble of mostly proprietary LLMs, aggregating their scores either by majority voting over binary ratings, or by averaging for ordinal scores. For pairwise evaluation, \citet{li2024prdpeerrankdiscussion} propose two methods which combine preferences of multiple LLMs, including an iterative multi-agent discussion.

\section{Problem Statement}
We formulate the problem of evaluating the output of an NLG system while providing error-based explanations as follows. Given an input $x$, an output $y$, and an evaluation aspect
$a$, the task is to return a tuple $\langle z, \{e_1, . . . , e_n\} \rangle$, where $z$ is a numeric score assigned to $y$, and $\{e_1, . . . , e_n\}$ represents a set of error annotations. Each error annotation $e_i = \langle s_i, t_i, l_i \rangle$ includes a span of text $s_i \in y$ corresponding to the problematic segment, a textual explanation $t_i$, and a severity level $l_i$. 
Although the term \emph{error} is used throughout this work, it should be understood in a broader sense as any issue in the text related to the evaluated aspect $a$. Examples of outputs provided by \textsc{OpeNLGauge} are presented in Figure~\ref{fig:exp1} and in Appendix \ref{appx:output_examples}.

\section{Open LLM Ensemble as Evaluator}
\label{sec:ensemble_evaluator}
To achieve a high-quality evaluation of NLG outputs, we propose \textsc{OpeNLGauge$_{ens}$}, a two-stage ensemble of open-weight LLMs. The ensemble consists of $n$ \emph{annotator models}, which perform independent analyses of the provided NLG output, and a \emph{consolidator model}, which is responsible for merging their results and filtering inaccuracies.
A high-level overview of the approach is presented in Figure~\ref{fig:ensemble_diagram}.

Although our approach requires multiple LLMs, using the ensemble with a handful of models ($n=5$ in our experiments) is still less computationally demanding than sampling multiple outputs to obtain statistical estimates, as required by some metrics, e.g.~G-Eval ($n=20$, \citet{liu-etal-2023-g}).
Furthermore, we only use quantized LLMs to limit the computational requirements.

\paragraph{Annotator models}

The annotator models are open-weight LLMs prompted to identify error spans in the text and to provide detailed explanations and severity levels for each error. This approach facilitates the interpretability of the evaluation process, but also acts as a chain-of-thought mechanism~\cite{10.5555/3600270.3602070}, helping the model to ground its decisions in a structured reasoning path. 

The full annotator model prompt is provided in Appendix~\ref{app:prompt_templates}. It contains the description of the evaluated task (e.g., data-to-text), the definition of the evaluated aspect, and a template for the model's response.
Since LLMs are known to confuse different evaluation aspects~\citep{hu-etal-2024-llm}, the prompt also contains several rules that instruct the model to remain focused on the specific evaluation aspect, not to make additional assumptions, and to justify any score lower than the maximum by at least one identified error.
Inspired by~\citet{liu-etal-2023-g}, we also include detailed steps for error identification. 

\label{sec:scoring-scales}
Furthermore, we provide a description of the overall scoring scale, including an explanation of the lowest and highest scores. The scale is presented as categorical (\emph{Unacceptable} < \emph{Poor} < \emph{Fair} < \emph{Good} < \emph{Excellent}), based on the intuition that adjectival categorical scales may be easier for language models to interpret than numerical scales. 
We use an integer severity scale (1-5) for scoring individual errors in order to avoid confusion with the overall scoring scale.\footnote{A categorical scale was also initially used for error severity. However, we found that LLMs sometimes confused the error severity scale with the overall score scale.}

Finally, the prompt contains the input that was originally used to generate the evaluated output (e.g., the source text for summarization or the input data in data-to-text). 
Some tasks may involve multiple inputs; for instance, evaluating knowledge-grounded question answering requires knowledge of both the question and the context. In such cases, the context is also presented to the model under separate headers.
Although structured formats like JSON might make parsing of the output easier, prompting LLMs to reason within strict structured outputs has been shown to impair their performance~\citep{tam-etal-2024-speak, kellner2024guiding}. Therefore, the models are instructed to produce textual outputs. 
\paragraph{Consolidator model}
The final score of the NLG output is computed as a simple average of the scores provided by the annotator models. However, to meet the requirement of explainable output, the error analyses of multiple annotators need to be unified. This is the task of the consolidator LLM.

This open-weight LLM is instructed to: (1) merge errors detected by multiple models, (2) unify their output format, and (3) clean up the annotations.
Specifically, the model is instructed to merge all error annotations that refer to the same issue at approximately the same location in the text, while maintaining annotation granularity.
It is also prompted to fix potential deviations from the expected output format. To simplify the cleaning process, error analyses produced by outlier annotator models\footnote{The annotation is considered to be an outlier if the score provided by the annotator differs from the mean score by at least two standard deviations, and this difference is at least 1.} are filtered out from the input to the consolidator model.
The full prompt for the consolidator model is provided in Appendix~\ref{app:prompt_templates}.

\section{Training a Distilled Model}

\subsection{Synthetic data generation}
\label{sec:dataset}
To distill knowledge from the ensemble, we collected the outputs of over a hundred NLG systems
and applied \textsc{OpeNLGauge$_{ens}$} to produce synthetic evaluation data.
We include NLG outputs produced on 15 datasets covering five task categories and almost 40 aspects. Table~\ref{tab:dataset_statistisc} shows the basic statistics of the constructed dataset.

\begin{table}[t]
\centering
\small
\begin{tabular}{>{\hspace{-1.5mm}}lcccc}
\toprule
\textbf{Task} & \textbf{Src.} & \textbf{Sys.} & \textbf{Asp.} & \textbf{Examples} \\
\midrule
Summarization & 5 & 39 & 6 & 12,070 \\ 
Data-to-text & 4 & 29 & 7 & \phantom{0}7,894 \\ 
Dialogue & 3 & 36 & 9 & 10,074 \\ 
Story Generation & 1 & 9 & 6 & \phantom{0}3,200 \\ 
Question Answering & 2 & 15 & 11 & \phantom{0}4,849 \\
\bottomrule
\end{tabular}
\caption{Training dataset statistics for \textsc{OpeNLGauge$_{ft}$}: number of source datasets (Src.), systems (Sys.), evaluation aspects (Asp.) and training examples for each task.}
\label{tab:dataset_statistisc}
\end{table}

The NLG outputs included cover the following tasks: data-to-text, summarization, question answering, dialogue response generation and story generation.
These were chosen to represent a diverse range of tasks, each with unique objectives, input-output relationships and evaluation aspects. Note that the underlying datasets are used only as inputs for NLG systems and do not include human evaluations of any kind.

For each task, we select a set of relevant evaluation aspects with their definitions. We consider two aspects distinct if they are associated with different tasks. For instance, \emph{coherence} in dialogue refers to coherence of the response with respect to the dialogue history, while in summarization it refers to the internal coherence of a summary. We list the datasets and aspects for each task in Appendix~\ref{appendix:datasets_aspects}.

To obtain output texts with varying quality and diverse types of errors, we sample outputs from a variety of systems, ranging from rule-based approaches to state-of-the-art LLMs. For older systems, we use pre-generated outputs from existing datasets, while outputs from more recent systems including LLMs are newly generated. An overview of the evaluated systems is shown in Appendix~\ref{appendix:system_outputs}.

To limit computational requirements for dataset generation, we apply a sampling procedure that ensures data diversity and broad coverage of different NLG systems and aspects while significantly reducing dataset size.
For each input, we randomly sample the outputs of $N$ systems, followed by sampling $M$ aspects for each input-output pair. The sampling includes all aspects described in Appendix~\ref{appendix:datasets_aspects} and all systems listed in Appendix~\ref{appendix:system_outputs}. This results in $N \, \times \, M$ (input, output, aspect) triples for each input, which are then passed to \textsc{OpeNLGauge$_{ens}$} to obtain synthetic annotations. For most tasks, we set $N = 4$ and $M$ = 3. 
This sampling strategy aims for a balanced distribution of inputs, NLG system outputs and evaluation aspects. It also ensures exposure to different outputs for the same input, i.e., it should not prime models to evaluate based solely on patterns in the inputs. Finally, by presenting multiple aspects for the same input-output pair, models are encouraged to learn differences in output quality between different evaluation aspects. 

To keep the merged evaluation outputs internally consistent, we remove outliers before merging, as described in Section~\ref{sec:ensemble_evaluator}. Table~\ref{tab:outliers} in Appendix~\ref{app:outliers} shows the proportions of outliers detected for all LLMs and task categories (3.4\% on average).

\subsection{Fine-tuning Procedure}
We use the dataset described in Section~\ref{sec:dataset} for supervised fine-tuning of a specialized LLM evaluator, with an instruction-tuned version of Llama 3.1 8B as the backbone. To avoid training the LLM to predict floating-point scores, we convert them to integers in the range 0-100 and then bin them to the nearest multiple of five. This extends the output space from five to twenty values (instead of 100), which is a trade-off between greater granularity in predictions and manageable task complexity.

As the model is expected to learn the task from training data, we used a simpler prompt template for fine-tuning, with a brief description of the task, the definition of the evaluated aspect, and the input and output to be evaluated (see Appendix~\ref{app:prompt_templates}).

\begin{table*}[!ht]
\small
\centering
\begin{tabular}{lccccccccc}
\toprule
\multirow{2}{*}{\textbf{Metric}} & \multicolumn{3}{c}{\textbf{QAGS-CNN/DM}} & \multicolumn{3}{c}{\textbf{QAGS-XSUM}} & \multicolumn{3}{c}{\textbf{Average}} \\
 & $r$ & $\rho$ & $\tau$ & $r$ & $\rho$ & $\tau$ & $r$ & $\rho$ & $\tau$ \\
\midrule
ROUGE-1 & 0.338 & 0.318 & 0.248 & -0.008 & -0.049 & -0.040 & 0.165 & 0.134 & 0.104 \\
ROUGE-2 & 0.459 & 0.418 & 0.333 & 0.097 & 0.083 & 0.068 & 0.278 & 0.250 & 0.200 \\
ROUGE-L & 0.357 & 0.324 & 0.254 & 0.024 & -0.011 & -0.009 & 0.190 & 0.156 & 0.122 \\
BERTScore & 0.576 & 0.505 & 0.399 & 0.024 & 0.008 & 0.006 & 0.300 & 0.256 & 0.202 \\
MoverScore & 0.414 & 0.347 & 0.271 & 0.054 & 0.044 & 0.036 & 0.234 & 0.195 & 0.153 \\
\midrule
FactCC & 0.416 & 0.484 & 0.376 & 0.297 & 0.259 & 0.212 & 0.356 & 0.371 & 0.294 \\
QAGS & 0.545 & - & - & 0.175 & - & - & 0.375 & - & - \\
BARTScore & 0.732 & 0.680 & 0.555 & 0.175 & 0.171 & 0.139 & 0.454 & 0.425 & 0.347 \\
UniEval & 0.682 & 0.662 & 0.532 & 0.461 & 0.488 & 0.399 & 0.572 & 0.575 & 0.466 \\
\midrule
G-Eval (GPT-3.5) & 0.477 & 0.516 & 0.410 & 0.211 & 0.406 & 0.343 & 0.344 & 0.461 & 0.377 \\
G-Eval (GPT-4) & 0.631 & 0.685 & 0.591 & 0.558 & 0.537 & 0.472 & 0.595 & 0.611 & 0.532 \\
LLM Evaluation (GPT-3.5) & 0.454 & 0.514 & 0.417 & 0.279 & 0.348 & 0.295 & 0.366 & 0.431 & 0.356 \\
LLM Evaluation (GPT-4) & 0.735 & 0.746 & 0.626 & 0.541 & 0.528 & 0.439 & 0.638 & 0.637 & 0.532 \\
\midrule
Auto-J & 0.291 & 0.238 & 0.214 & 0.225 & 0.214 & 0.203 & 0.258 & 0.226 & 0.209 \\
TIGERScore & 0.574 & 0.562 & 0.479 & 0.424 & 0.445 & 0.412 & 0.499 & 0.504 & 0.446 \\
InstructScore & 0.287 & 0.278 & 0.233 & -0.096 & -0.134 & -0.119 & 0.095 & 0.072 & 0.057 \\
Themis & \textbf{0.747} & \textbf{0.761} & \textbf{0.680} & \underline{0.599} & 0.607 & \underline{0.546} & \underline{0.673} & \underline{0.684} & \textbf{0.613} \\
\midrule
\textsc{OpeNLGauge$_{ens}$} & \underline{0.738} & \underline{0.753} & 0.627 & \textbf{0.630} & \textbf{0.624} & 0.531 & \textbf{0.684} & \textbf{0.689} & 0.579 \\
 \textbullet \space Command R+ 104B & 0.676 & 0.675 & 0.617 & 0.540 & 0.541 & 0.515 & 0.608 & 0.608 & 0.566 \\
 \textbullet \space Gemma 2 27B & 0.579 & 0.646 & 0.579 & 0.592 & \underline{0.614} & \textbf{0.563} & 0.585 & 0.630 & 0.571 \\
 \textbullet \space Llama 3.1 Nemotron 70B & 0.705 & 0.733 & \underline{0.650} & 0.587 & 0.586 & 0.540 & 0.646 & 0.659 & \underline{0.595} \\
 \textbullet \space Mistral Large 2 123B & 0.658 & 0.704 & 0.635 & 0.577 & 0.570 & 0.541 & 0.617 & 0.637 & 0.588 \\
 \textbullet \space Qwen 2.5 72B & 0.678 & 0.720 & 0.635 & 0.568 & 0.569 & 0.526 & 0.623 & 0.644 & 0.581 \\
\midrule
Llama 3.1 8B & 0.275 & 0.242 & 0.219 & 0.218 & 0.230 & 0.218 & 0.247 & 0.236 & 0.219 \\
\textsc{OpeNLGauge$_{ft}$} & 0.668 & 0.695 & 0.584 & 0.607 & 0.607 & 0.524 & 0.638 & 0.651 & 0.554 \\
\bottomrule
\end{tabular}
\caption{Segment-level Pearson ($r$), Spearman ($\rho$) and Kendall ($\tau$) correlations of different metrics for factual consistency on QAGS. The best correlations are highlighted in bold, the second best are underlined.}
\label{tab:meta_eval_qags}
\end{table*}

\section{Experimental Setup}

\subsection{Ensemble}

The ensemble consists of six open-weight LLMs: five annotators and one consolidator. Each selected LLM is distributed under a license that permits at least non-commercial use and allows the model's outputs to be used as training data. At the time of the experiments, these models ranked among the top-performing open-weight LLMs on the Chatbot Arena Leaderboard\footnote{\url{https://lmarena.ai/leaderboard}} 
\citep{chiang2024chatbot}. 

The annotator models include the following: Llama 3.1 Nemotron 70B \citep{wang2024helpsteer2}, Qwen 2.5 72B \citep{yang2024qwen2}, Gemma 2 27B \citep{gemma2024gemma}, Command R+ 104B \citep{cohere_for_ai_2024}, and Mistral Large 2 123B\footnote{\url{https://mistral.ai/news/mistral-large-2407/}}.
We apply Llama 3.3 70B \citep{dubey2024llama} as the consolidator model. To address computational constraints, we use quantized versions available through the Ollama platform.\footnote{\url{https://ollama.com/}}
For synthetic data generation, we set the temperature to zero to obtain consistent results. For details on the models, see Appendix~\ref{app:llms}.

\subsection{Distillation}

To produce \textsc{OpeNLGauge$_{ft}$}, we apply LoRA~\cite{hu2022lora} with rank 16 and alpha 32 to fine-tune the instruction-tuned version of Llama 3.1 8B. The model is trained for one epoch with a learning rate of 2e-4, using AdamW optimizer \citep{loshchilov2017weight} with a weight decay of 0.01. We apply a linear learning rate schedule with a warm-up period corresponding to the first 5\% of training steps. The batch size is set to 16. This setup enables resource-efficient training that requires only around 20GB of VRAM, completing one training epoch in just six hours on a single A40 GPU.

\subsection{Evaluation Datasets}

We used seven popular meta-evaluation datasets to assess how our metric correlates with human judgments (for detailed descriptions of these datasets, including details on aggregation of human scores, see Appendix~\ref{app:metaeval}).
The datasets cover the following tasks: summarization -- SummEval \citep{fabbri2021summeval}, QAGS \citep{wang2020asking}; story generation -- HANNA \citep{chhun2022human}; data-to-text -- SFRES and SFHOT \citep{wen2015semantically}, text simplification -- Wiki-DA \citep{alva2021suitability} and dialogue generation -- TopicalChat \citep{gopalakrishnan2023topical}. For \textsc{OpeNLGauge$_{ft}$}, text simplification is a task unseen during training and TopicalChat contains an unseen aspect (groundedness).
For human evaluation of error spans, we used the RotoWire dataset~\citep{thomson2020gold} with annotations from a data-to-text task in the basketball domain, a task unseen during training by the evaluated fine-tuned metrics.

\begin{table}[t]
\centering\small
\begin{tabular}{>{\hspace{-1.5mm}}l>{\hspace{-1.5mm}}c>{\hspace{-1.5mm}}c>{\hspace{-1.5mm}}c}
\toprule
\textbf{Dataset} & \textbf{Llama 3.1} & \textbf{OpeNLGauge$_{ft}$} & $\Delta$ \\
\midrule
QAGS & 0.236 & 0.651 & \textcolor{increase_green}{+0.415} \\ 
SummEval & 0.186 & 0.502 & \textcolor{increase_green}{+0.316} \\ 
TopicalChat & 0.309 & 0.578 & \textcolor{increase_green}{+0.269} \\ 
SFRES/SFHOT & 0.108 & 0.315 & \textcolor{increase_green}{+0.207} \\ 
HANNA & 0.150 & 0.425 & \textcolor{increase_green}{+0.275} \\
Wiki-DA & 0.405 & 0.789 & \textcolor{increase_green}{+0.384} \\
\bottomrule
\end{tabular}
\caption{Comparison of Spearman ($\rho$) correlations of the backbone model (Llama 3.1 8B) and our metric \textsc{OpeNLGauge$_{ft}$} fine-tuned from the backbone on the dataset described in Section \ref{sec:dataset}. For each dataset, the correlations are averaged across all evaluated aspects.}
\label{tab:llama_improvement}
\end{table}

\subsection{Baselines}

We compare our methods with a variety of commonly used evaluation metrics, including traditional metrics such as ROUGE \citep{lin-2004-rouge}, BLEU \citep{papineni2002bleu} and METEOR \citep{agarwal2008meteor}, distance-based metrics MoverScore \citep{zhao-etal-2019-moverscore} and BERTScore \citep{zhang2019bertscore}, and trained metrics BARTScore \citep{yuan2021bartscore} and UniEval \citep{zhong2022towards}. 
We also include the following LLM-based metrics: GPTScore \citep{fu2024gptscore}, G-Eval \citep{liu-etal-2023-g}, LLM Evaluation \citep{chiang-lee-2023-closer}, Prometheus~\citep{kim2023prometheus}, Auto-J~\citep{li2023generative}, InstructScore~\citep{xu-etal-2023-instructscore}, TIGERScore~\citep{jiang2023tigerscore} and Themis~\citep{hu2024themis}. To measure the improvement of \textsc{OpeNLGauge$_{ft}$} relative to the base model, we include the instruction-tuned version of Llama 3.1 8B as an additional baseline. 
The specific metrics reported differ depending on the dataset and the evaluated NLG task.

Additionally, our comparisons include some task-specific and aspect-specific metrics: QAGS \citep{wang2020asking} and FactCC \citep{kryscinski-etal-2020-evaluating} for evaluating factual consistency in summarization, SARI \citep{xu-etal-2016-optimizing} and LENS \citep{maddela2022lens} for text simplification, and USR \citep{mehri2020usr} for dialogue response generation tasks. The latter metric has several variants; we use the variant with the best Pearson correlation for each aspect in our comparison. 

\section{Results}

\begin{figure}[t]
    \centering
    \includegraphics[width=\columnwidth]{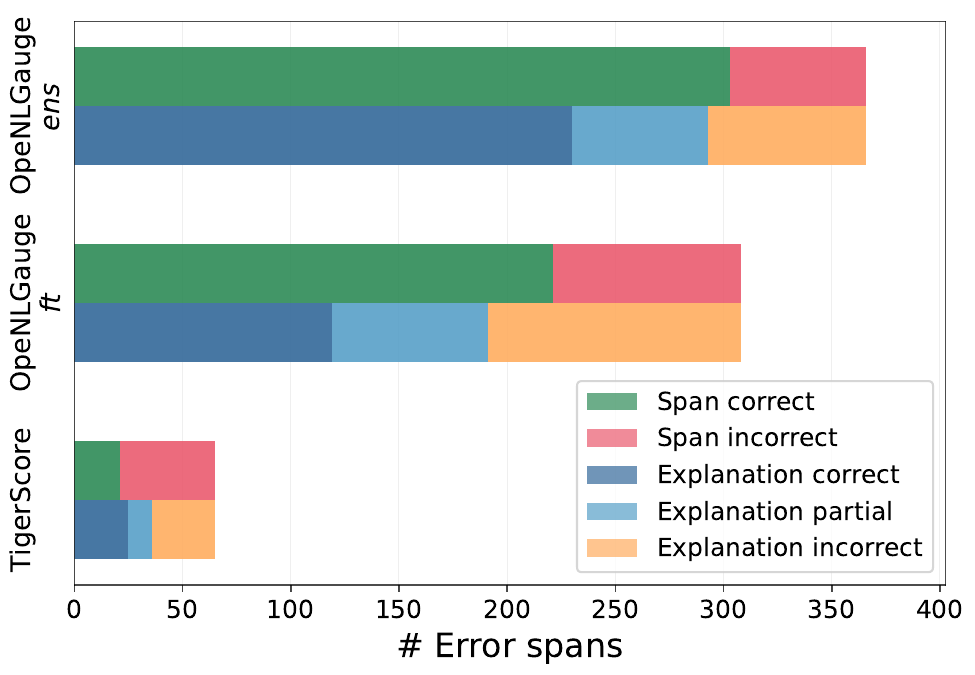}
    \caption{Results of human evaluation of error spans and explanations. Top half of each bar: Error spans marked as \textcolor{span_correct}{correct} or \textcolor{span_incorrect}{incorrect} (hallucinated spans, no span provided, or spans without errors). Bottom half: Explanations marked as \textcolor{explanation_correct}{correct}, \textcolor{explanation_partial}{partial} (partially correct or incomplete) or  \textcolor{explanation_incorrect}{incorrect} (not addressing actual errors, vague or incorrect). The differences between TigerScore and \textsc{OpeNLGauge$_{ens}$} are statistically significant (t-test, $p < 0.05$). See Table~\ref{tab:openlgauge} for more details.}
    \label{fig:human_eval}
\end{figure}

\begin{figure*}[t]
    \centering
    \includegraphics[width=0.9\textwidth]{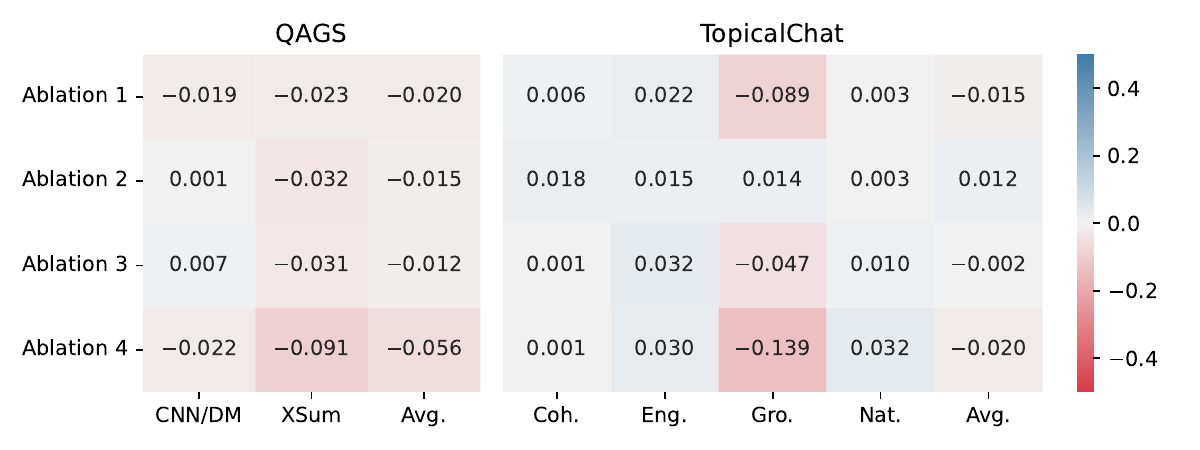}
    \caption{Ablation results on QAGS and TopicalChat for \textsc{OpeNLGauge$_{ens}$}. Plotted values represent differences in Spearman's $\rho$ correlations with human scores between the ensemble with the original prompt and the corresponding ablation. For TopicalChat, \textbf{Coh.} = coherence, \textbf{Eng.} = engagingness, \textbf{Gro.} = groundedness, \textbf{Nat.} = naturalness, \textbf{Avg.} = average for all aspects.}
    \label{fig:ablation}
\end{figure*}

\subsection{Score Correlation with Humans}

The results for factual consistency on QAGS are presented in Table~\ref{tab:meta_eval_qags} and the results for other datasets are available in Tables~\ref{tab:meta_eval_summeval}--\ref{tab:meta_eval_wikida} in Appendix~\ref{appx:full_results}. On QAGS, \textsc{OpeNLGauge$_{ens}$} achieves the highest average performance on both Pearson's $r$ and Spearman's~$\rho$. On Kendall's $\tau$, our method is outperformed by Themis, which can be attributed to different score granularities used by these two methods (see Section~\ref{sec:kendall_correlation} for a discussion).
The distilled version of our metric was consistently the third best measure. Notably, it outperformed the metrics based on the proprietary GPT-4 on this task.

On SummEval (Table~\ref{tab:meta_eval_summeval}), the best performing metric was Themis, closely followed by \textsc{OpeNLGauge$_{ens}$}. However, note that training data for Themis include almost 62,000 \emph{human-annotated} examples for the summarization task.
Comparing our approach to TigerScore, another method that provides a similar level of explainability (error span annotations), we observe 15 p.p.\ improvements on Spearman's $\rho$ averaged over all aspects.
The smaller \textsc{OpeNLGauge$_{ft}$} outperformed all other fine-tuned LLM-based metrics except Themis, and also surpassed metrics based on prompting GPT-3.5 by a large margin.

On TopicalChat (Table~\ref{tab:meta_eval_topical_chat}), the \emph{LLM Evaluation} metric based on GPT-4 emerged as the strongest evaluator, while \textsc{OpeNLGauge$_{ens}$} ranked between this method and its GPT-3.5-based version. \textsc{OpeNLGauge$_{ft}$} achieved 27 p.p.\ improvement in Spearman's $\rho$ compared to its Llama 3.1 backbone.

In data-to-text, \textsc{OpeNLGauge$_{ens}$} excelled in evaluating naturalness, while \textsc{OpeNLGauge$_{ft}$} stands out as the strongest evaluator of informativeness on the SFRES dataset (Table \ref{tab:meta_eval_sfres_sfhot}). Interestingly, on data-to-text problems, the distilled version of our metric achieved slightly better results on average than our ensemble. A similar situation is observed for story generation (Tables \ref{tab:meta_eval_hanna_r}--\ref{tab:meta_eval_hanna_tau}), where the distilled version obtained better average scores on Spearman's $\rho$ and Kendall's $\tau$, and closely followed \textsc{OpeNLGauge$_{ens}$} on Pearson's $r$.

In text simplification (Table \ref{tab:meta_eval_wikida}), our ensemble achieved superior performance across the board, outperforming LENS, a strong baseline metric specialized for this task.

Additionally, Table~\ref{tab:llama_improvement} presents a comparison of \textsc{OpeNLGauge$_{ft}$} with Llama 3.1 8B (instruct), indicating a considerable improvement over the prompted model.

\paragraph{Generalization to unseen tasks} Although not originally trained on text simplification, \textsc{OpeNLGauge$_{ft}$} outperforms all baseline metrics and individual LLMs in averaged Pearson correlation on the Wiki-DA dataset (Table \ref{tab:meta_eval_wikida}) and achieves a particularly high score on meaning preservation.

\paragraph{Generalization to unseen aspects} On TopicalChat (Table~\ref{tab:meta_eval_topical_chat}), a large improvement for \textsc{OpeNLGauge$_{ft}$} over Llama 3.1 8B is observed on groundedness, which is an aspect unseen during training. \textsc{OpeNLGauge$_{ft}$} also surpassed most fine-tuned metrics and most individual LLMs on this aspect. Moreover, Wiki-DA contains additional unseen aspects (meaning preservation and simplicity) on which \textsc{OpeNLGauge$_{ft}$} shows considerable improvement over Llama 3.1 8B and outperforms most of the individual larger LLMs.

\subsection{Human Evaluation of Error Spans}
\label{sec:human-eval}

We performed a small in-house human evaluation study to compare the quality of explanations obtained by \textsc{OpeNLGauge} and TigerScore, another LLM-based metric that also provides error-span annotations (see Table~\ref{tab:m-comparision}).
For this purpose, we used a data-to-text task in the basketball domain \citep{thomson2020gold}. Five expert annotators evaluated the output of all three systems on 40 instances (a total of 120 outputs and 950 error spans).
We asked the annotators to:
(1) evaluate provided error spans, marking them as \emph{correctly identified}, \emph{not containing an error}, \emph{hallucinated}, and situations where \emph{no span was provided};
(2) evaluate generated explanations, marking them as \emph{correct}, \emph{partially correct}, \emph{incomplete}, \emph{vague}, \emph{incorrect}, or texts that \emph{do not describe an error}.

To assess reliability of our human annotation, we computed Cohen's $\kappa$ coefficient \citep{cohen1960coefficient} of inter-annotator agreement on 50 error spans with double annotations. We obtained $\kappa = 0.82$ for the evaluation of error spans, and $\kappa = 0.46$ for error explanations.

The results are shown in Figure \ref{fig:human_eval}, with more details provided in Table~\ref{tab:openlgauge} in Appendix~\ref{appx:full_results}. Both \textsc{OpeNLGauge$_{ft}$} and \textsc{OpeNLGauge$_{ens}$} are over twice more accurate than TigerScore at annotating error spans, while finding over ten times more correct error spans.
The task of providing accurate error explanations was more difficult for all the approaches evaluated.  
\textsc{OpeNLGauge$_{ens}$} achieved the highest performance and was almost twice as accurate as TigerScore and \textsc{OpeNLGauge$_{ft}$}, which achieved similar accuracies.

\subsection{Ablation Experiments}

\paragraph{Prompt ablations}
We explore the effect of using different scales for the overall score and error severity: integer scales for both (Ablation 1), integer scale for overall score and categorical scale for severity (Ablation 2), and categorical scale for both (Ablation 3). Recall that \textsc{OpeNLGauge} uses a categorical scale for overall score and integer scale for error severity -- see Section~\ref{sec:scoring-scales}.

Correlation differences between the full prompt and the ablations are summarized in Figure~\ref{fig:ablation}, for detailed results see Appendix~\ref{app:ablations}. Overall, the change of scale has little effect on the average correlation of the whole ensemble, but it has a dramatic effect on some individual LLMs and aspects. This illustrates the ability of the ensemble to compensate for weaknesses in individual annotator models.

Finally, we examine the effect of removing the evaluation rules from the prompt (Ablation 4), which has an inconsistent effect on different models/aspects, but on average degrades the ensemble evaluation quality -- up to 5.6 p.p.

\paragraph{Ensemble structure}
We also analyze the effect of ensemble size and the influence of its particular components on the correlations with human scores by recomputing the results for all ensemble combinations. The results are presented in Appendix~\ref{app:ablations2}. For Wiki-DA, performance increases with ensemble size, with the full ensemble being the best combination. For other datasets, there are a few smaller combinations that actually rank higher, but none is consistently better than the full ensemble.

\paragraph{Inter-annotator agreement between LLMs}
To obtain additional insights into how the individual models of the ensemble diverge in their overall score predictions, we compute several measures of inter-annotator agreement. The results presented in Appendix~\ref{appx:llm_iaa} indicate only low to moderate agreement for most of the datasets, especially on exact overall scores, which suggests a sufficient diversity for combining outputs of these models into an ensemble.

\paragraph{Error analysis aggregation}
The consolidator model aggregates error span annotations from multiple LLMs, which could potentially lead to an overall larger number of detected errors. To estimate the extent of over-annotation by both the ensemble and its components, we analyze the number of detected errors for output-aspect pairs rated with maximum score by human annotators. The results presented in Appendix~\ref{app:false_positives} indeed show some tendency of the ensemble to over-annotate due to error accumulation from the individual models. However, most spans annotated by the ensemble were marked as correct in the experiment in Section~\ref{sec:human-eval}. This could indicate that OpeNLGauge$_{ens}$ finds subtle errors which human annotators overlook, but further analysis of this discrepancy is needed.

\paragraph{Score aggregation}
Additionally, we compare different methods of aggregating overall scores of individual LLMs to a final score in Appendix \ref{appx:aggregation_methods}. These results indicate that despite its simplicity, simple averaging is the most effective approach, generally providing the highest correlations with human scores.

\section{Summary}
In this work, we present \textsc{OpeNLGauge} -- a versatile method for evaluating NLG that uses only open-weight models and provides fine-grained explainability. The method provides a much better explanation quality than previous methods and achieves competitive correlations with human judgments.

\section*{Limitations}

\textsc{OpeNLGauge} is a method for evaluating a variety of NLG tasks. While this paper presents the evaluation on several NLG tasks and the method achieves good performance on unseen aspects, domains and tasks, the actual performance on new NLG tasks is unknown.
In particular, the metric has not been tested in a multilingual setting. 
Moreover, previous research has shown that some LLM-based metrics have a bias towards texts generated by LLMs~\cite{liu-etal-2023-g}. 

\section*{Acknowledgments}

This work was supported by the European Research Council (Grant agreement No.~101039303, NG-NLG) and the National Recovery Plan funded project MPO 60273/24/21300/21000 CEDMO 2.0 NPO. It used resources of the LINDAT/\hspace{0mm}CLARIAH-CZ Research Infrastructure (Czech Ministry of Education, Youth, and Sports project No. LM2018101).

\bibliography{custom}

\clearpage

\appendix

\section{Aspects and datasets in the training data}
\label{appendix:datasets_aspects}

This section provides a detailed overview of all source datasets and evaluation aspects used to generate our training data for \textsc{OpeNLGauge$_{ft}$}.

\subsection{Summarization}

As source data for summarization, we utilize five datasets spanning three distinct tasks: news article summarization, forum post summarization and dialogue summarization. Our evaluation dataset includes six commonly used aspects: four related to meaning of the evaluated text, one to its form and one to both. Some of these aspects overlap significantly in their definitions. This is intentional, as our goal is to make the evaluator model robust to small variations in aspect definitions, and enable generalization to new aspects. Table \ref{tab:system_outputs_summ} provides an overview of all summarization systems in our dataset.

\subsubsection{Datasets}

\begin{itemize}
    \item \textbf{CNN/DailyMail} \citep{hermann2015teaching} is a popular summarization dataset that consists of news articles from CNN and DailyMail, paired with corresponding bullet-point summaries. Originally developed for question answering, it was later adapted for summarization \citep{nallapati2016abstractive}. For this dataset, we use pre-generated outputs from \citet{stiennon2020learning}, which include results from 11 different systems, including human references and an extractive baseline. Since our meta-evaluation datasets for summarization contain inputs from CNN/DailyMail, we ensure there is no overlap in the source texts between these datasets when sampling the inputs.
    \item \textbf{Newsroom} \citep{grusky2018newsroom} is a large-scale dataset of news articles and their summaries, collected from diverse sources, domains, authors and time range. In addition to newly generated LLM outputs, we use outputs of the systems evaluated in the original paper, which include three summarization systems and two extractive baselines. 
    \item \textbf{SAMSum} dataset \citep{gliwa2019samsum} addresses the dialogue summarization task, containing dialogues with short summaries created by linguists. In addition to outputs from several different LLMs, we use pre-generated outputs of six systems from \citep{gao2022dialsummeval}. 
    \item \textbf{TL;DR} \citep{volske2017tl} is a collection of Reddit posts with user-created summaries. Unlike many popular datasets that focus on news articles, TL;DR contains informal and less structured texts spanning diverse topics. Similarly to CNN/DailyMail, we use outputs from \citet{stiennon2020learning}.
    \item \textbf{XSum} \citep{narayan2018don} is a dataset for the \textit{extreme summarization} task, designed for abstractive approaches. Unlike datasets such as Newsroom or CNN/DailyMail, which favor extractive summarization, XSum contains BBC articles paired with concise, single-sentence summaries. We utilize pre-generated outputs from five systems and baselines evaluated in the original paper, along with newly generated output from LLMs.
\end{itemize}

\subsubsection{Aspects}

\begin{itemize}
    \item \textbf{Consistency} evaluates whether the summary is factually aligned with the source text. This involves determining if the facts in the summary can be entailed by the source. Consistency is closely tied to hallucinations, which may be categorized as factual or non-factual. In our approach, all information is required to be supported by the source text, making both types of hallucinations inconsistent with the source. The definition of consistency used in our prompts is: \textit{Extent to which the facts in the summary are consistent with the source text. Factually consistent summary should not contain facts that are not supported by the source text.}
    \item \textbf{Accuracy} is largely synonymous with consistency. It evaluates whether the factual information from the source is accurately represented in the summary. Following \citet{stiennon2020learning}, we define accuracy as: \textit{Extent to which the factual information in the summary accurately matches the source text. An accurate summary should not contain information that is not present in the source text, should not contradict the source text, and generally should not be misleading.}
    \item \textbf{Relevance} of a summary is concerned with content selection. A relevant summary should include important points from the source text while omitting unimportant details. Compared to consistency, relevance is more subjective, as determining what information should or should not be included in a summary can sometimes be ambiguous. The definition used in our evaluation is: \textit{Extent to which the summary captures important information of the source text. A relevant summary should include all and only important information from the source text.}
    \item \textbf{Coverage} evaluates how much of the important information from the source text is covered by the summary. In this sense, it is closely related to relevance, however, it does not include non-redundancy as a criterion. We use a definition adapted from \citet{stiennon2020learning}: \textit{Extent to which the summary covers the important information in the source text. A summary has good coverage if it mentions the main information from the source text that is important to understand the events described in the text. A summary has poor coverage if someone reading only the summary would be missing several important pieces of information about the event in the source text.}
    \item \textbf{Coherence} refers to the structural quality of the summary and involves attributes such as cohesion, consistency and relevance \citep{reinhart1980conditions}. It is determined by both semantic and formal structure of the text. We define coherence as: \textit{Extent to which the summary is well-structured and organized, presenting information in a logical order that flows naturally from sentence to sentence. Coherent summary forms a unified body of information and makes it easy to understand the main ideas.}
    \item \textbf{Fluency} focuses on the formal quality of the text, including grammaticality and naturalness. Unlike coherence, fluency is concerned with sentence–level quality rather than the overall structure of the text. We define fluency as: \textit{Formal quality of individual sentences of the summary. A fluent sentence should be grammatical, natural and easy to understand.}
\end{itemize}

\begin{table*}[t]
\centering
\resizebox{\textwidth}{!}{
\begin{tabular}{lll}
\toprule
\textbf{Format} & \textbf{Example} & \textbf{Datasets} \\
\midrule
JSON & \texttt{\{"name": "The Phoenix", "eatType": "pub", "food": "Indian", ...\}} & E2E NLG \\ 
attribute-value (1) & \texttt{name: The Phoenix\textcolor{purple}{\textbackslash{}n}eatType: restaurant\textcolor{purple}{\textbackslash{}n}food: Indian\textcolor{purple}{\textbackslash{}n} ...} & E2E NLG \\ 
attribute-value (2) & \texttt{name[The Phoenix], eatType[restaurant], food[Indian], ...} & E2E NLG \\ 
RDF (1) & \texttt{"The\_Velvet\_Underground | genre | Proto-punk" ...} & WebNLG 2020 \\ 
RDF (2) & \texttt{(The\_Velvet\_Underground, genre, Proto-punk) ...} & WebNLG 2020 \\
CSV & \texttt{united states,32,1,31,12\textcolor{purple}{\textbackslash{}n}australia,5,0,5,3\textcolor{purple}{\textbackslash{}n} ...} & LogicNLG \\ 
linearized table & \texttt{<page\_title> List of Norwegian fjords </page\_title> <section\_title> ...} & ToTTo \\ 
\bottomrule
\end{tabular}
}
\caption{Input data formats used in our dataset for data-to-text tasks.}
\label{tab:data_formats}
\end{table*}

\subsection{Data-to-text}

For data-to-text category, we collected inputs from four datasets that represent four distinct tasks: table-to-text, RDF-to-text, attribute-value list to text and logical NLG. Since it is important for an NLG evaluation method to reliably evaluate outputs with respect to structured data in different formats, we include a number of different input formats in our training data, including JSON, CSV and linearized tables with markup. Table \ref{tab:data_formats} provides an overview of the input formats. The list of evaluated systems in our dataset is shown in Table \ref{tab:system_outputs_d2t}.

\subsubsection{Datasets}

\begin{itemize}
    \item \textbf{E2E NLG} dataset \citep{duvsek2020evaluating, novikova2017e2e} was chosen as a representative of simple data-to-text tasks. In this dataset, models are tasked with generating descriptions of restaurant venues based on attribute-value-based meaning representations (MRs). Target descriptions were crowdsourced using textual and pictorial representations of MRs as stimuli. To ensure diversity, we selected pre-generated outputs from five systems described by \citep{duvsek2020evaluating}, considering their model architectures and evaluation results across various metrics and aspects. These were extended by new outputs from several LLMs. The inputs and pre-generated outputs were sampled from the test set. Three different input formats are used in the training data (see Table \ref{tab:data_formats}).
    \item \textbf{WebNLG 2020} \citep{ferreira20202020} is an RDF-to-text dataset designed for generating natural language text from RDF triples collected from DBPedia knowledge base \citep{mendes2012dbpedia}. Each input contains between one and seven triples, where each triple represents a binary relation in the form (subject, property, object). Similarly to E2E NLG, we selected a diverse set of output systems based on model architectures and evaluation results from the WebNLG+ 2020 Challenge, and used additional LLMs to generate new outputs.
    \item \textbf{ToTTo} \citep{parikh2020totto} is an open-domain table-to-text generation dataset. It consists of Wikipedia tables with highlighted cells, and the task is to generate single-sentence descriptions of the data in these highlighted cells. Inputs are provided in two formats: full tables, which include indices to highlighted cells, and linearized tables, where only the highlighted data is presented in a linear order, while the structure is annotated with markup tags. As we observed that even medium-sized open-weight LLMs often struggle with hallucinations when using full tables, we restricted our evaluation to linearized tables.
    \item \textbf{LogicNLG} \citep{chen2020logical} introduced the task of logical NLG, where models generate statements that can be logically entailed from the data in a table. This task involves various aggregations and comparisons, making it more difficult than simply transforming structured data to free-form text. Although the task explicitly requires generating five logical statements per input, we sampled between one to five statements from each generated output to increase the diversity of output lengths. Inputs were formatted as CSV with “|” as a separator. In addition to new LLM outputs, we used existing system outputs from \citet{zhao-etal-2023-investigating}.
\end{itemize}

\subsubsection{Aspects}

\begin{itemize}
    \item \textbf{Faithfulness} measures whether all information in the generated text is supported by the data, making it equivalent to precision. Similarly to factual consistency in summarization, we consider both factual and non-factual hallucinations as errors. For our evaluations, we define faithfulness as: \textit{Extent to which the information in the text is supported by the data.}
    \item \textbf{Correctness} evaluates whether the information from the data is accurately presented in the generated text. The output is maximally correct if it does not contain any incorrect statements with respect to the input. Correctness overlaps significantly with faithfulness, but its definition varies based on the specific task. For example, we define correctness for LogicNLG as: \textit{Extent to which the statements are logically and factually correct with respect to the provided data.}
    \item \textbf{Coverage} refers to the degree to which the generated text covers the information in the data. The output has maximum coverage when all information from the data is included in the text, which makes it analogous to recall. For example, in WebNLG 2020, it evaluates whether all predicates and their arguments are mentioned, while in ToTTo, it determines whether all highlighted table cells are described. We apply coverage to evaluate all datasets except LogicNLG, where the task is to infer interesting observations, rather than fully cover the source data. We define coverage as follows, with slight variations depending on the task: \textit{Extent to which the text includes description of all information presented in the data.}
    \item \textbf{Informativeness} is closely related to coverage, as it evaluates how much of the information the generated text provides. However, it does not require complete coverage of the data, therefore it is applicable to tasks such as LogicNLG, where full coverage of a table is not necessary. The definition used in our evaluation depends on the particular dataset. For example, the definition used for LogicNLG is: \textit{Extent to which the statements provide interesting or useful information about the data.}
    \item \textbf{Fluency} refers to the formal quality of the generated text, and includes grammaticality, naturalness and readability. Some definitions also include coherence \citep{ferreira20202020}, although coherence is usually treated as a separate aspect. For our evaluation, we define fluency as: \textit{Extent to which the text is grammatical, natural and easy to understand.}
    \item \textbf{Grammaticality} focuses on the correctness of grammar and spelling in the generated text. A text is fully grammatical if it contains no grammatical or spelling errors. While grammaticality is often included as a sub-aspect of fluency, both aspects are commonly used in practice. Therefore, we include it to help the model learn differences between evaluation aspects on different levels of hierarchy. In our dataset, grammaticality is defined as: \textit{Extent to which the text is grammatical (free of grammar and spelling errors).}
    \item \textbf{Naturalness} refers either to the human-likeness of the text, or the likelihood that it was produced by a native speaker. Like grammaticality, naturalness is often treated as a component of fluency. Additionally,  its evaluation often includes assessment of grammaticality, as this can often be an indicator of whether the text was produced by a native speaker. This illustrates how evaluation aspects often overlap or have hierarchical relationships. For our purposes, naturalness is defined as: \textit{Extent to which the text is likely to have been produced by a native speaker.}
\end{itemize}
    
\subsection{Dialogue Response Generation}

For dialogue response generation, we source the inputs from three dialogue datasets, focusing on open-domain non-task-oriented dialogue response generation. Table \ref{tab:system_outputs_dialogue} provides an overview of all evaluated systems in the training dataset.

\subsubsection{Datasets} 

\begin{itemize}
    \item \textbf{Wizard of Wikipedia} \citep{dinan2018wizard} consists of conversations grounded in one of 1365 topics and corresponding knowledge retrieved from Wikipedia. In these conversations, either participant may select the topic and initiate the discussion, although they have asymmetric roles. One participant takes on the role of the wizard, an expert with access to a topic-relevant knowledge, on which they can base their responses. The other participant acts as an apprentice, a curious learner that is eager to discuss the chosen topic. To create inputs of varying length, we randomly select a dialogue history length between two turns and the full conversation, and truncate the dialogue to this length. The last utterance is replaced by a system output, except when the reference is used as evaluated output. In addition to newly generated LLM responses, we include human responses from the original dataset in the outputs.
    \item \textbf{EmpatheticDialogues} \citep{rashkin2018towards} is a dataset of dialogues grounded in emotional situations, designed to train and evaluate dialogue models on empathetic response generation. Each conversation is associated with an emotional label, where one of the participants describes a situation in which they experienced a given emotion. We sample up to five turns from each dialogue and replace the last utterance with a generated response. The emotion label is used as additional context for annotator LLMs to evaluate the appropriateness and empathy of the responses, but is excluded from the prompts used for system output generation.
    \item \textbf{DailyDialog} \citep{li2017dailydialog} includes conversations on various daily life topics, annotated with emotion labels and communicative intents. We include data from DailyDialog to represent diverse topics and scenarios in the training set. Alongside newly generated responses, we also collect pre-generated outputs from three sources \citep{gupta2019investigating, huang2020grade, zhao2020designing} to represent older dialogue systems.  
\end{itemize}

\subsubsection{Aspects}

\begin{itemize}
    \item \textbf{Coherence} in dialogue is a concept slightly different from coherence in tasks that involve generation of standalone texts, such as summarization or story generation. In dialogue, it measures how meaningful and logically consistent the response is with the preceding conversation. This includes not only alignment of the response with the last utterance, but also consistency with the dialogue participant's earlier responses in terms of logic and style. In our evaluation, coherence is broadly defined as: \textit{Extent to which the response is a meaningful continuation of previous dialogue.}
    \item \textbf{Relevance} evaluates how closely a response aligns with the topic of conversation. In this sense, this aspect overlaps with coherence to some degree. We define relevance as: \textit{Extent to which the response is relevant and on-topic given the dialogue history.}
    \item \textbf{Appropriateness} addresses whether the response is semantically and pragmatically appropriate in the given context. Depending on the definitions, appropriateness might overlap to a large extent with coherence and relevance. For our evaluation, we define appropriateness as: \textit{Extent to which the response is semantically and pragmatically appropriate given the conversation history.}
    \item \textbf{Empathy} is evaluated specifically on responses from the EmpatheticDialogues dataset, where the goal is to determine whether the response acknowledges and reflects the emotions of the other participant. We define empathy as: \textit{Extent to which the response shows understanding of the feelings of the person talking about their experience.}
    \item \textbf{Interestingness} is concerned with the informational value of the response, specifically whether it presents stimulating ideas, facts or opinions. As it is one of the more subjective evaluation aspects, we define it vaguely and let the annotator models determine the criteria for interestingness: \textit{Extent to which the response is interesting given the dialogue history.}
    \item \textbf{Engagingness} is closely related to interestingness but is sometimes treated as a distinct aspect (e.g., \citealp{mehri2020unsupervised, see-etal-2019-makes}). While interestingness focuses on the context itself, engagingness emphasizes maintaining the user's attention and encouraging them to continue with the conversation. For our purposes, engagingness is defined as: \textit{Extent to which the response captures and maintains the user's interest, encouraging further interaction. Engaging responses contain opinions, preferences, thoughts or interesting facts.}
    \item \textbf{Fluency} in dialogue response generation has a similar meaning to its use in other tasks and refers to the formal quality of the response. We define fluency as: \textit{Extent to which the response is grammatically correct, natural and fluent.}
    \item \textbf{Understandability} evaluates both the content and form of a response, focusing on its clarity and ease of comprehension. The definition we use for our evaluation is: \textit{Extent to which the response is easy to understand and comprehend given the dialogue history.}
\end{itemize}

\subsection{Story Generation}
\label{sec:story_generation}

The NLG tasks discussed so far generally contain inputs that are relatively longer compared to the outputs. This pattern is especially common in tasks like summarization and dialogue generation, although certain data-to-text tasks also share this characteristic. To represent scenarios with short inputs and long outputs, we include story generation in the training data. Table \ref{tab:system_outputs_story} lists all evaluated systems in our dataset. 

\subsubsection{Datasets}

As the source of inputs, we use \textbf{WritingPrompts} \citep{fan2018hierarchical}, a story generation dataset derived from Reddit's WritingPrompts subreddit, where users submit prompts that can inspire other users to write stories. The dataset consists of a diverse range of topics, story lengths and writing styles. We reuse existing outputs from the OpenMEVA dataset \citep{guan2021openmeva} and generate additional outputs by four LLMs. To increase the diversity of generated stories in terms of their length, we generate the outputs with two different prompt versions, each requiring a different length of the story. The outputs are then randomly sampled from either the shorter or the longer set. As we observed a tendency of LLMs to generate a long list of errors for longer inputs, our evaluator models are instructed to limit the number of identified errors to a maximum of eight, and to prioritize the most severe ones if necessary.

\subsubsection{Aspects}

While relevance and coherence are two commonly used aspects for story generation, there is no consensus on which other evaluation aspects are the most relevant. Inspired by social sciences, \citet{chhun2022human} propose four additional aspects, aimed at providing a complete and non-redundant set of criteria. Following their work, we adopt the aspects defined in the HANNA benchmark, using our own definitions for most of them:

\begin{itemize}
    \item \textbf{Relevance} measures the degree to which a story aligns with the given prompt \citep{chhun2022human, chiang2023can}, title \citep{jhamtani2020narrative, yao2019plan, xie2023next} or story beginning \citet{wang2020narrative}. In some cases, relevance also evaluates whether the story remains on-topic for its duration (e.g., \citealp{goldfarb2020content}). Since our inputs are prompts, we define relevance as: \textit{Extent to which the story is relevant to the writing prompt.}
    \item \textbf{Coherence} in story generation typically refers to logical consistency and narrative flow (e.g., \citealp{yao2019plan, li2023collaborative, jhamtani2020narrative}). Other works define coherence more vaguely, such as how much the story ``makes sense'' \citep{chhun2022human} or how well its sentences ``fit together'' \citep{xie2023next}. For our purposes, coherence is defined as: \textit{Extent to which the story is logically consistent and coherent.}
    \item \textbf{Engagement} is a subjective and often vaguely defined aspect that evaluates how engaging the story is to the reader (e.g., \citealp{chhun2022human, li2023collaborative}). Due to its inherent subjectivity, we apply a simple definition and leave its interpretation to the evaluator models: \textit{Extent to which the story is engaging and interesting.}
    \item \textbf{Empathy} is related to emotional commentary and empathy, and refers to how well the story conveys character's emotions. We define empathy as: \textit{The clarity and depth with which the character's emotions are conveyed in the story.}
    \item \textbf{Surprise} is concerned with the story's ending, and evaluates its unexpectedness and originality. We define surprise as: \textit{How surprising the end of the story was.}
    \item \textbf{Complexity} measures how intricate and elaborate the story is. Complexity is not necessarily an aspect of quality, but rather a feature of the text. Whether greater complexity is desired or not depends on the audience. Our definition of complexity is: \textit{How elaborate the story is.}
\end{itemize}

\subsection{Question Answering}

The question answering subset of our dataset consists of two distinct tasks: narrative question answering and table question answering. The inputs consist of a question and structured data in which the answer should be grounded. The evaluated systems in our training data are listed in Table \ref{tab:system_outputs_qa}.

\subsubsection{Datasets}

\begin{itemize}
    \item \textbf{NarrativeQA} \citep{kovcisky2018narrativeqa} consists of human-written questions and free-form answers based on stories or their summaries. The stories include books and movie scripts and are provided either as full texts or as human-written summaries. Since full stories do not fit into the context window of many LLMs, we use only summaries to generate the outputs. Along with newly generated answers, we include human reference answers from the original datasets in the evaluated outputs.
    \item \textbf{FeTaQA} \citep{nan2022fetaqa} is a question answering dataset based on Wikipedia tables that requires models to aggregate and reason about the entities in the table and their relations. This places the task at the intersection of data-to-text and question answering task categories. In addition to new LLM outputs, we use pre-generated outputs from \citet{zhao-etal-2023-investigating}.
\end{itemize}

\subsubsection{Aspects}

\begin{itemize}
    \item \textbf{Correctness} evaluates if the answer to a question is correct with respect to the input. Since our models are instructed to assess the quality on an ordinal scale, we evaluate a \emph{degree} of correctness -- the answer should receive the maximum score if it is fully correct, while lower scores should reflect the number and severity of correctness issues. We define correctness as: \textit{Extent to which the answer to the question is correct with respect to the input.}
    \item \textbf{Informativeness} addresses whether all information required by the question is provided in the answer. We define informativeness as: \textit{Extent to which the answer provides all information that the question asked for.}
    \item \textbf{Completeness} evaluates comprehensiveness of the answer and the degree to which all aspects of the question are covered. The meaning is slightly different from informativeness, which is concerned with the information that the question explicitly asks for. In our dataset, completeness is defined as: \textit{Extent to which the answer is comprehensive and ensures all question aspects are addressed.}
    \item \textbf{Conciseness} measures the degree to which an answer is focused and directly answers the question without unnecessary details and elaboration. Although the goal might often be to generate both complete and concise answers, these two aspects might correlate negatively. Conciseness is defined as: \textit{Extent to which the answer is concise and to the point.}
    \item \textbf{Relevance} is concerned with the specificity of an answer and measures the degree to which the answer addresses the particular question asked. Although it is related to conciseness, relevance is not that much concerned with the amount of detail in the answer. We define relevance as: \textit{Extent to which the answer is specific and meaningful with respect to the question.}
    \item \textbf{Factuality} evaluates factual consistency of the answer with the provided context. In our dataset, this aspect is used in the narrative question answering task, and we use a similar definition as in summarization: \textit{Extent to which the answer is supported by the summary.}
    \item \textbf{Faithfulness} is used for the table question answering task and is synonymous with factuality. We define faithfulness as: \textit{Extent to which the information presented in the answer is supported by the input.}
    \item \textbf{Fluency} in question answering refers to the formal quality of the answer and is defined similarly as in the other tasks presented so far: \textit{Extent to which the response is grammatical, natural and easy to understand.}
    \item \textbf{Naturalness} is interpreted in the same way as in data-to-text and dialogue response generation tasks, and the definition we apply is: \textit{Extent to which the answer is likely to have been produced by a native speaker.}
    \item \textbf{Grammaticality} measures the grammatical quality of the answer and is defined as: \textit{Extent to which the answer is grammatical (free of grammar and spelling errors)}. 
\end{itemize}

\section{Collected system outputs}
\label{appendix:system_outputs}

Tables \ref{tab:system_outputs_summ}--\ref{tab:system_outputs_qa} list the evaluated systems for each task category in our dataset. 

\begin{table*}[htbp]
\centering
\small
\caption{Overview of the evaluated systems in the dataset for the summarization task. Sources: GW = \citet{gao2022dialsummeval}, GR = \citet{grusky2018newsroom}, NA = \citet{narayan2018don}, ST = \citet{stiennon2020learning}, new = newly generated.}
\begin{tabular}{@{}lll@{}}
\toprule
\textbf{System} & \textbf{Type} & \textbf{Sources} \\
\midrule
Qwen 2.5 0.5B \citep{yang2024qwen2} & Instruction-tuned LLM & new \\
Llama 2 7B Chat \citep{touvron2023llama} & Instruction-tuned LLM & new \\
Gemma 2 2B \citep{gemma2024gemma} & Instruction-tuned LLM & new \\
Nous Hermes 2 Mixtral 8x7B DPO\footnote{\url{https://huggingface.co/NousResearch/Nous-Hermes-2-Mixtral-8x7B-DPO}} & Instruction-tuned LLM & new \\
GPT-4o\footnote{\url{https://openai.com/index/hello-gpt-4o/}} & Instruction-tuned LLM & new \\

\midrule
OpenAI summarization (pre-trained) & pre-trained LMs & ST \\
OpenAI summarization (supervised) & LMs trained with SFT & ST \\
OpenAI summarization (RLHF) & LMs trained with SFT+PPO & ST \\
T5 \citep{raffel2020exploring} & pre-trained LM & ST \\
UniLM \citep{dong2019unified} & pre-trained LM & ST \\
CODS \citep{wu-etal-2021-controllable} & BART-based hybrid model & GW \\
ConvoSumm \citep{fabbri-etal-2021-convosumm} & BART-based model & GW \\
Ctrl-DiaSumm \citep{liu-chen-2021-controllable} & BART-based model & GW \\

\midrule
ConvS2S \citep{gehring2017convolutional} & Convolutional seq-to-seq & NA \\
Topic-ConvS2S \citep{narayan2018don} & Topic-conditioned ConvS2S & NA \\
S2S \citep{cho2014learning, sutskever2014sequence} & RNN-based seq-to-seq with attention & GR \\
PNG \citep{vinyals2015pointer, gu̇lccehre2016pointing} & Pointer-generator network & GR, GW, NA \\
TextRank \citep{barrios2016variations} & Ranking-based extractive summarization & GR \\

\midrule
Reference & Human-written reference & ST \\
Title & Extractive baseline & ST \\
LEAD & Extractive baseline & NA \\
LEAD-2 & Extractive baseline & ST \\
LEAD-3 \citep{see2017get} & Extractive baseline & GR, GW, ST \\
Ext-Oracle & Extractive oracle & NA \\
Fragments & Extractive oracle & NA \\
\bottomrule
\end{tabular}
\label{tab:system_outputs_summ}
\end{table*}

\begin{table*}[htbp]
\centering
\small
\caption{Overview of the evaluated systems in the dataset for the data-to-text task. \textbf{Sources}: DU = \citet{duvsek2020evaluating}, FE = \citet{ferreira20202020}, ZH = \citet{zhao-etal-2023-investigating}, new = newly generated.}
\begin{tabular}{@{}lll@{}}
\toprule
\textbf{System} & \textbf{Type} & \textbf{Sources} \\
\midrule
Qwen 2.5 Coder 1.5B \citep{yang2024qwen2} & Instruction-tuned LLM & new \\
Gemma 2 2B \citep{gemma2024gemma} & Instruction-tuned LLM & new \\
Llama 2 7B Chat \citep{touvron2023llama} & Instruction-tuned LLM & new \\
Solar 10.7B \citep{kim-etal-2024-solar} & Instruction-tuned LLM & new \\
DeepSeek Coder v2 16B \citep{zhu2024deepseek} & Instruction-tuned LLM & new \\
Nous Hermes 2 Mixtral 8x7B DPO\footnote{\url{https://huggingface.co/NousResearch/Nous-Hermes-2-Mixtral-8x7B-DPO}} & Instruction-tuned LLM & new \\
Claude 3.5 Sonnet\footnote{\url{https://www.anthropic.com/news/claude-3-5-sonnet}} & Instruction-tuned LLM & new \\
GPT-4o\footnote{\url{https://openai.com/index/hello-gpt-4o/}} & Instruction-tuned LLM & new \\
\midrule

NILC \citep{sobrevilla-cabezudo-pardo-2020-nilc-webnlg} & Fine-tuned BART & FE \\
Orange-NLG \citep{montella2020denoising} & Fine-tuned BART & FE \\
Amazaon AI (Shanghai) \citep{guo-etal-2020-2} & Graph CNN + T5 & FE \\
GPT2-C2F \citep{chen2020logical} & Fine-tuned GPT-2 & ZH \\
LoFT \citep{zhao2023loft} & Fine-tuned BART & ZH \\
PLOG \citep{liu-etal-2022-plog} & Fine-tuned T5 & ZH \\
R2D2 \citep{nan2022r2d2} & Fine-tuned T5 & ZH \\
Flan-T5 \citep{chung2024scaling} & Instruction-tuned LM & ZH \\
\midrule

Adapt \citep{elder-etal-2018-e2e} & RNN seq-to-seq & DU \\
Sheff2 \citep{chen2018sheffield} & RNN seq-to-seq & DU \\
Slug \citep{juraska-etal-2018-deep} & RNN + convolutional seq-to-seq & DU \\
\midrule

Forge1 \citep{mille2018forge} & Rule-based & DU \\
TR2 \citep{smiley-etal-2018-e2e} & Template-based & DU \\
DANGNT-SGU \citep{tran-nguyen-2020-webnlg} & Template-based & FE \\
RALI-Université de Montréal \citep{lapalme-2020-rdfjsrealb} & Template-based & FE \\

\bottomrule
\end{tabular}
\label{tab:system_outputs_d2t}
\end{table*}

\begin{table*}[htbp]
\centering
\small
\caption{Overview of the evaluated systems in the dataset for the dialogue response generation task. Sources: GU = \citet{gupta2019investigating}, HU = \citet{huang2020grade}, ZH = \citet{zhao2020designing}, new = newly generated.}
\begin{tabular}{@{}lll@{}}
\toprule
\textbf{System} & \textbf{Type} & \textbf{Sources} \\
\midrule
Claude 3.5 Sonnet\footnote{\url{https://www.anthropic.com/news/claude-3-5-sonnet}} & Instruction-tuned LLM & new \\
GPT-4o\footnote{\url{https://openai.com/index/hello-gpt-4o/}} & Instruction-tuned LLM & new \\
Tülu 3 \citep{lambert2024tulu} & Instruction-tuned LLM & new \\
Dolphin 2.9 Llama 3 8B\footnote{\url{https://huggingface.co/cognitivecomputations/dolphin-2.9-llama3-8b}} & Instruction-tuned LLM & new \\
Vicuna 7B \citep{zheng2023judging} & Instruction-tuned LLM & new \\

\midrule
BlenderBot-small \citep{roller-etal-2021-recipes} & Dialogue LM & new \\
DialoGPT-small \citep{zhang-etal-2020-dialogpt} & Dialogue LM & new \\
GPT-Neo 125M \citep{gao2020pile} & Pre-trained LM & new \\
GPT-2 \citep{wolf2019transfertransfo} & Pre-trained LM & ZH \\

\midrule
CVAE \citep{zhao2017learning} & Conditional variational autoencoder & GU \\
HRED \citep{serban2016building} & Hierarchical recurrent encoder-decoder & GU, ZH \\
Transformer-generator \citep{dinan2018wizard} & Transformer-based generative model & HU \\
Transformer-ranker \citep{urbanek-etal-2019-learning} & Transformer-based ranking model & HU \\
DualEncoder \citep{lowe2015ubuntu} & LSTM dual encoder & GU \\
VHRED \citep{serban2017hierarchical} & Latent variable HRED & ZH \\
S2S \citep{cho2014learning, sutskever2014sequence} & RNN-based seq-to-seq with attention & GU, ZH \\
\bottomrule
\end{tabular}
\label{tab:system_outputs_dialogue}
\end{table*}

\begin{table*}[htbp]
\centering
\small
\caption{Overview of the evaluated systems in the dataset for the story generation task. Sources: GU = \citet{guan2021openmeva}, new = newly generated.}
\begin{tabular}{@{}lll@{}}
\toprule
\textbf{System} & \textbf{Type} & \textbf{Sources} \\
\midrule
Gemma 2 2B \citep{gemma2024gemma} & Instruction-tuned LLM & new \\
Dolphin 2.9 Llama 3 8B\footnote{\url{https://huggingface.co/cognitivecomputations/dolphin-2.9-llama3-8b}} & Instruction-tuned LLM & new \\
Nous Hermes 2 Mixtral 8x7B DPO\footnote{\url{https://huggingface.co/NousResearch/Nous-Hermes-2-Mixtral-8x7B-DPO}} & Instruction-tuned LLM & new \\
GPT-4o\footnote{\url{https://openai.com/index/hello-gpt-4o/}} & Instruction-tuned LLM & new \\

\midrule
GPT-2 \citep{radford2019language} & Pre-trained LM & GU \\
GPT-KG \citep{guan2020knowledge} & Knowledge-enhanced GPT-2 & GU \\

\midrule
Fusion \citep{fan2018hierarchical} & Convolutional seq-to-seq with attention & GU \\
Plan\&Write \citep{yao2019plan} & Hierarchical RNN-based model & GU \\
S2S \citep{cho2014learning, sutskever2014sequence} & RNN-based seq-to-seq & GU \\

\bottomrule
\end{tabular}
\label{tab:system_outputs_story}
\end{table*}

\begin{table*}[htbp]
\centering
\small
\caption{Overview of the evaluated systems in the dataset for the question answering task. Sources: KO = \citet{kovcisky2018narrativeqa}, ZH = \citet{zhao-etal-2023-investigating}, new = newly generated.}
\begin{tabular}{@{}lll@{}}
\toprule
\textbf{System} & \textbf{Type} & \textbf{Sources} \\
\midrule
Claude 3.5 Sonnet\footnote{\url{https://www.anthropic.com/news/claude-3-5-sonnet}} & Instruction-tuned LLM & new \\
GPT-4o\footnote{\url{https://openai.com/index/hello-gpt-4o/}} & Instruction-tuned LLM & new \\
Nous Hermes 2 Mixtral 8x7B DPO\footnote{\url{https://huggingface.co/NousResearch/Nous-Hermes-2-Mixtral-8x7B-DPO}} & Instruction-tuned LLM & new \\
DeepSeek Coder v2 16B \citep{zhu2024deepseek} & Instruction-tuned LLM & new \\
Llama 2 7B Chat \citep{touvron2023llama} & Instruction-tuned LLM & new \\
Qwen 2.5 Coder 1.5B \citep{yang2024qwen2} & Instruction-tuned LLM & new \\
Qwen 2.5 0.5B \citep{yang2024qwen2} & Instruction-tuned LLM & new \\

\midrule
GPT-Neo 125M \citep{gao2020pile} & Pre-trained LM & new \\
BART \citep{lewis2019bart} & Pre-trained LM & ZH \\
Flan-T5 \citep{chung2024scaling} & Instruction-tuned LM & ZH \\
OmniTab \citep{jiang-etal-2022-omnitab} & Table pre-trained LM & ZH \\
ReasTAP \citep{zhao-etal-2022-reastap} & Table pre-trained LM & ZH \\
TAPEX \citep{liu2022tapex} & Table pre-trained LM & ZH \\

\midrule
Reference & Human-written reference & KO \\

\bottomrule
\end{tabular}
\label{tab:system_outputs_qa}
\end{table*}

\section{LLMs used in the experiments}
\label{app:llms}
The specific versions of LLMs used in the experiments are presented in Table~\ref{tab:ollama_quantization}.
\begin{table*}[ht]
\small
    \centering
    \begin{tabular}{lcl}
        \toprule
        \textbf{Model} & \textbf{Quantization} & \textbf{Tag} \\
        \midrule
        Command R+ 104B & 5-bit & \texttt{command-r-plus:104b-08-2024-q5\_K\_M} \\
        Gemma 2 27B & 8-bit & \texttt{gemma2:27b-instruct-q8\_0} \\
        Llama 3.1 Nemotron 70B & 8-bit & \texttt{nemotron:70b-instruct-q8\_0} \\
        Llama 3.3 70B & 8-bit & \texttt{llama3.3:70b-instruct-q8\_0} \\
        Mistral Large 2 123B & 4-bit & \texttt{mistral-large:123b-instruct-2407-q4\_K\_M} \\
        Qwen 2.5 72B & 8-bit & \texttt{qwen2.5:72b-instruct-q8\_0} \\
        \bottomrule
    \end{tabular}
    \caption{Quantization levels and Ollama tags used for the models.}
\label{tab:ollama_quantization}
\end{table*}
%--------------------------

\section{Outliers}
\label{app:outliers}

Table \ref{tab:outliers} shows the percentages of outliers detected for each annotator model in the ensemble and task category during synthetic data generation. To keep the merged evaluation outputs internally consistent, these outliers were removed before merging the outputs of the individual LLMs.

The proportions in the table indicate that the final annotation utilized most of the generated evaluations from the individual LLMs. The exceptions include Command R+ 104B and Gemma 2 27B on data-to-text tasks, with 29\% and 10\% outliers, respectively. Additionally, 10\% evaluation outputs of Command R+ 104B were removed for question answering tasks. However, on average only 3.4\% of evaluation outputs are detected as outliers across all model-task pairs.

\begin{table*}[ht]
\centering
\small
\begin{tabular}{lccccc}
\toprule
\textbf{Model} & \textbf{Data-to-text} & \textbf{Summarization} & \textbf{Dialogue} & \textbf{Story Generation} & \textbf{Question Answering} \\
\midrule
Command R+ 104B & 28.88 & 0.13 & 0.73 & 0.94 & 9.61 \\
Gemma 2 27B & 10.32 & 1.96 & 1.52 & 1.03 & 5.07 \\
Llama 3.1 Nemotron 70B &  4.27 & 1.48 & 4.02 & 1.03 & 0.85 \\
Mistral Large 2 123B & 4.89 & 0.93 & 0.63 & 0.25 & 1.22 \\
Qwen 2.5 72B & 1.39 & 1.36 & 1.43 & 1.12 & 0.47 \\
\bottomrule
\end{tabular}
\caption{Proportions (\%) of evaluation outputs detected as outliers for each model and task category. Across all model-task pairs, 3.4\% of evaluation outputs are detected as outliers on average (median = 1.36\%).}
\label{tab:outliers}
\end{table*}

\section{Meta-evaluation datasets}
\label{app:metaeval}
We used the following datasets for meta-evaluation. In the datasets where human scores are not already aggregated, we averaged the scores of all annotators.
\begin{itemize}
    \item \textbf{SummEval} \citep{fabbri2021summeval} is a standard meta-evaluation dataset for summarization. It consists of summaries generated from CNN/DailyMail articles, with 100 input articles and 16 different system outputs for each article. Human evaluations address four aspects: \textit{(factual) consistency}, \textit{relevance}, \textit{coherence}, and \textit{fluency}. Each output is scored by three expert annotators and five crowdworkers. We use only expert scores to measure correlations.
    \item \textbf{QAGS} \citep{wang2020asking} contains annotations of \textit{consistency} for summaries from CNN/DailyMail and XSum datasets. It includes 235 CNN/DailyMail summaries and 239 XSum summaries, each annotated by three evaluators. Annotators assign a binary factual consistency score (yes/no) for each sentence of the summary. We follow \citep{wang2020asking} and apply majority voting for each sentence annotation, followed by averaging sentence-level scores to obtain the overall score for the summary.
\item \textbf{HANNA} \citep{chhun2022human} is a benchmark for story generation based on the WritingPrompts dataset. It contains three human scores for each of the 1056 stories across six aspects: \textit{relevance}, \textit{coherence}, \textit{engagement}, \textit{empathy}, \textit{surprise} and \textit{complexity} (see Appendix~\ref{sec:story_generation} for details on these aspects).
\item \textbf{SFRES} and \textbf{SFHOT} \citep{wen2015semantically} are used to perform a meta-evaluation for data-to-text. These datasets consist of dialogue acts (DAs) in structured format with generated responses providing information about restaurants and hotels in San Francisco. Human judgments are provided for \textit{informativeness} and \textit{naturalness}.
    \item \textbf{TopicalChat} \citep{gopalakrishnan2023topical} annotations from the USR dataset \citep{mehri2020usr} are used for meta-evaluation of dialogue response generation.
 The dataset includes human evaluations for five aspects: \textit{groundedness}, \textit{coherence}, \textit{interestingness}, \textit{naturalness} and \textit{understandability}.
  As the source data differ from our training data, TopicalChat serves as an out-of-domain evaluation dataset. Additionally, since \textit{groundedness} is not present in our training data, we evaluate it as an unseen aspect. 
\item \textbf{Wiki-DA} \citep{alva2021suitability} is a dataset of DA human ratings for text simplification, a task unseen by \textsc{OpeNLGauge$_{ft}$} during training. Along with scores for \textit{fluency}, this dataset also includes two unseen aspects: \textit{meaning preservation} and \textit{simplicity}.
\end{itemize}

\section{Prompt templates}
\label{app:prompt_templates}

Prompt templates for annotator and consolidator LLMs are presented in Listings~\ref{lst:prompt-annotator}--\ref{lst:prompt-supervisor}. Prompt template used for fine-tuning and inference of the distilled model is shown in Listing~\ref{lst:prompt-train}.

\section{Full results}

\begin{table}[t]
    \centering \small
    \begin{tabular}{ll>{\hspace{-3mm}}rrrrr}
    \toprule
    \multirow{2}{*}{\rotatebox{90}{\bf System}}& \multirow{2}{*}{\rotatebox{90}{\raisebox{-2.8cm}{\textbf{Error span}}}} & \multirow{2}{*}{\rotatebox{90}{Error}} & \multirow{2}{*}{\rotatebox{90}{No span given}}  & \multirow{2}{*}{\rotatebox{90}{Not an error}} & \multirow{2}{*}{\rotatebox{90}{Hallucination}} & \multirow{2}{*}{\rotatebox{90}{Total}} \\
     & \rule{0pt}{1.4cm}\textbf{Explanation}  &  &  &  &  &  \\
    \midrule
    \multirow{9}{*}{\rotatebox{90}{OpeNLGauge\textsubscript{ens}}} & Correct & 218 & 9 & 3 & 0 & \shade 230\\
     & Partially correct & 41 & 0 & 0 & 1 & \shade 42 \\
     & Incomplete & 20 & 0 & 1 & 0 & \shade 21 \\
     & Vague & 4 & 2 & 1 & 0 & \shade 7 \\
     & Incorrect & 20 & 3 & 32 & 1 & \shade 56 \\
     & Not an error & 0 & 0 & 10 & 0 & \shade 10 \\
     & \shade  Total & \shade 303 & \shade 14 & \shade  47 & \shade  2 & \shade  366 \\
     \cmidrule(lr){2-7}
     & \emph{Span OK (\%)} & \textbf{83} & & & & \\
     & \emph{Exp. correct (\%)} & \textbf{63} & & & & \\
    \midrule
    \multirow{9}{*}{\rotatebox{90}{OpeNLGauge\textsubscript{ft}}} & Correct & 108 & 3 & 4 & 4 & \shade 119 \\
     & Partially correct & 60 & 1 & 2 & 0 & \shade 63 \\
     & Incomplete & 9 & 0 & 0 & 0 & \shade 9 \\
     & Vague & 3 & 9 & 0 & 0 & \shade 12 \\
     & Incorrect & 41 & 7 & 42 & 6 & \shade 96 \\
     & Not an error & 0 & 1 & 8 & 0 & \shade 9 \\
     & \shade Total & \shade 221 & \shade 21 & \shade 56 & \shade 10 & \shade 308 \\
     \cmidrule(lr){2-7}
     & \emph{Span OK (\%)} & \textbf{72} & & & & \\
     & \emph{Exp. correct (\%)} & \textbf{39} & & & & \\
    \midrule
    \multirow{9}{*}{\rotatebox{90}{TigerScore}} & Correct & 8 & 17 & 0 & 0 & \shade 25 \\
     & Partially correct & 5 & 2 & 0 & 0 & \shade 7 \\
     & Incomplete & 2 & 2 & 0 & 0 & \shade 4 \\
     & Vague & 0 & 3 & 0 & 0 & \shade 3 \\
     & Incorrect & 5 & 16 & 2 & 0 & \shade 23 \\
     & Not an error & 1 & 2 & 0 & 0 & \shade 3 \\
     & \shade Total & \shade 21 & \shade 42 & \shade 2 & \shade 0 & \shade 65 \\
     \cmidrule(lr){2-7}
     & \emph{Span OK (\%)} & \textbf{32} & & & & \\
     & \emph{Exp. correct (\%)} & \textbf{38} & & & & \\
    \bottomrule
    \end{tabular}
    \caption{Detailed results of human evaluation of error-spans and their explanations. Each table shows absolute occurrence counts of different error span and explanation validity (see Section~\ref{sec:human-eval}), with overall proportions of correct annotation given below. The reported differences between TigerScore and \textsc{OpeNLGauge$_{ens}$} are statistically significant (t-test, $p < 0.05$).}
    \label{tab:openlgauge}
\end{table}

\label{appx:full_results}
Tables \ref{tab:meta_eval_summeval}--\ref{tab:meta_eval_wikida}
contain the meta-evaluation results for additional datasets described in Appendix \ref{app:metaeval}. Detailed results of human evaluation of error span quality are presented in Table \ref{tab:openlgauge}.

\subsection{Kendall's $\tau$ correlations}
\label{sec:kendall_correlation}

On QAGS, our method achieves lower Kendall's $\tau$ than Themis, although it surpasses the metric on Spearman's $\rho$. This discrepancy can be attributed to a difference in score precisions between these two methods. As a result of averaging, \textsc{OpeNLGauge$_{ens}$} provides more granular floating point scores, while Themis predicts integer scores on a Likert scale. Generally, we can expect more tied pairs (i.e. pairs that are neither concordant nor discordant) in the calculation of Kendall's $\tau$ with integer scores, which can have a substantial effect on the correlations. In QAGS, 78\% of human scores map to integer scores after normalization. Therefore, we consider Spearman's $\rho$ more appropriate for comparing evaluation capabilities of different metrics.

Similarly, specific individual LLMs score higher on Kendall's $\tau$ than the ensemble on some datasets (see Tables \ref{tab:meta_eval_qags}, \ref{tab:meta_eval_summeval}, \ref{tab:meta_eval_hanna_rho} and \ref{tab:meta_eval_hanna_tau}). This is due to the same reason outlined above. Specifically, there are 37\% tied pairs for \textsc{OpeNLGauge$_{ens}$} and human scores, while the individual LLMs, which provide integer scores, show substantially larger proportions of tied pairs: between 46\% and 54\%.

\section{Ablation experiments}
\subsection{Prompt ablations}
\label{app:ablations}
\begin{figure*}[h!]
    \centering
    \includegraphics[width=\textwidth]{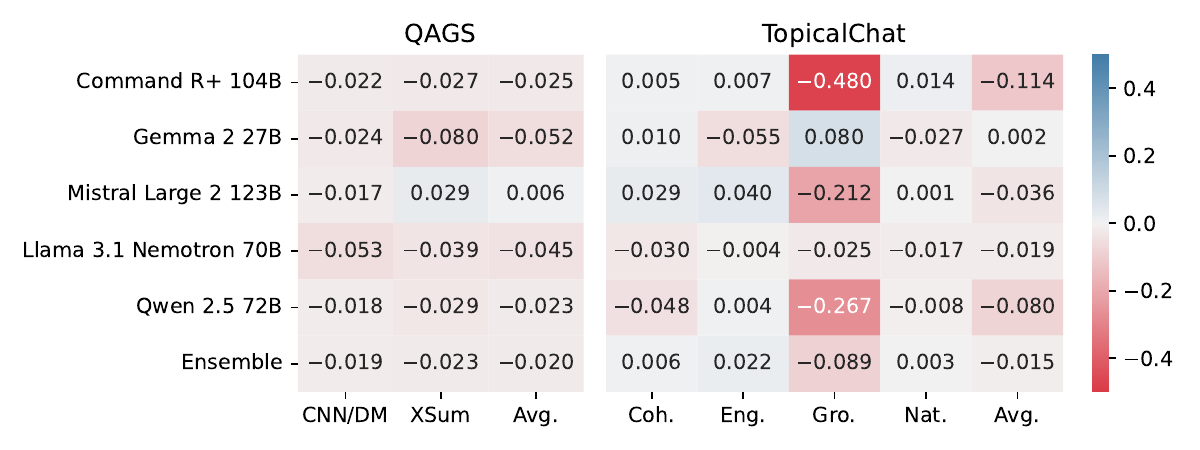}
    \caption{Results for Ablation 1 on QAGS and TopicalChat. The LLMs are instructed to provide both integer overall scores (1--5) and integer severity levels (1--5). The plotted values represent differences in Spearman's $\rho$ correlations with human scores between the original prompt and the ablation. For TopicalChat, \textbf{Coh.} = coherence, \textbf{Eng.} = engagingness, \textbf{Gro.} = groundedness, \textbf{Nat.} = naturalness, \textbf{Avg.} = average of all aspects.}
    \label{fig:ablation1}
\end{figure*}

\begin{figure*}[h!]
    \centering
    \includegraphics[width=\textwidth]{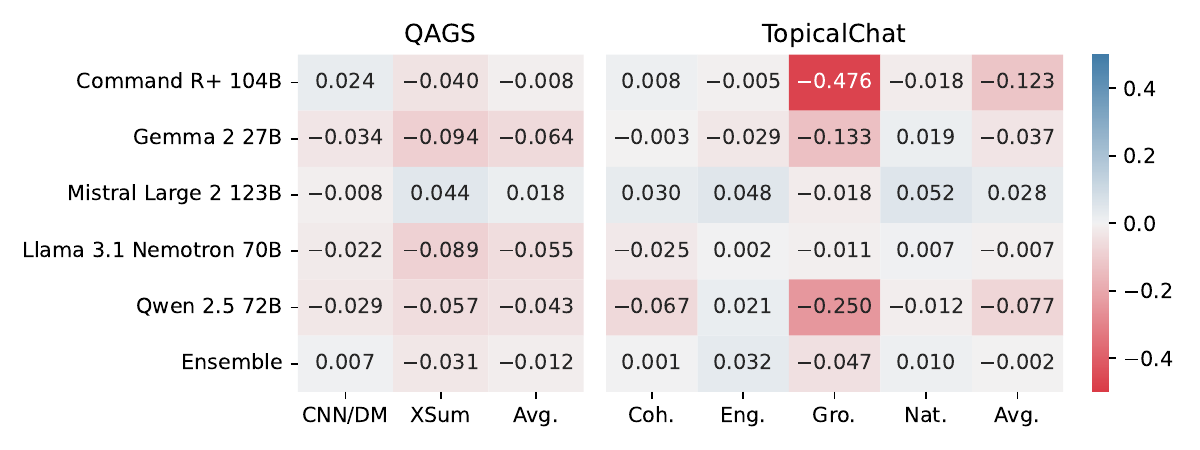}
    \caption{Results for Ablation 2 on QAGS and TopicalChat. The LLMs are instructed to provide integer overall scores (1--5), and categorical severity levels on the following scale: \emph{Neutral}, \emph{Minor}, \emph{Moderate}, \emph{Major}, \emph{Critical}.}
    \label{fig:ablation2}
\end{figure*}

\begin{figure*}[h!]
    \centering
    \includegraphics[width=\textwidth]{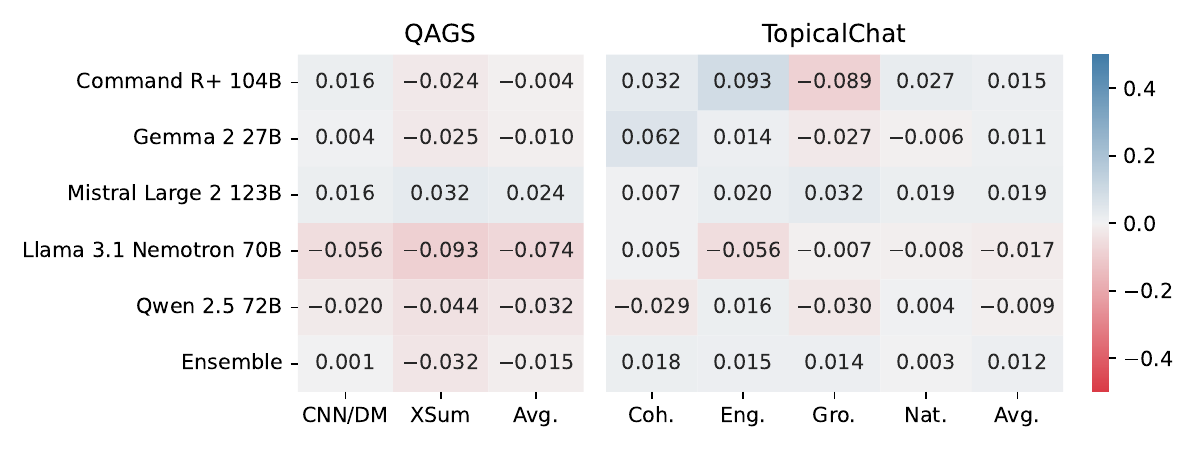}
    \caption{Results for Ablation 3 on QAGS and TopicalChat. The LLMs are instructed to provide categorical overall scores on the scale described in Section \ref{sec:ensemble_evaluator}, and categorical severity levels on the following scale: \emph{Neutral}, \emph{Minor}, \emph{Moderate}, \emph{Major}, \emph{Critical}.}
    \label{fig:ablation3}
\end{figure*}

\begin{figure*}[h!]
    \centering
    \includegraphics[width=\textwidth]{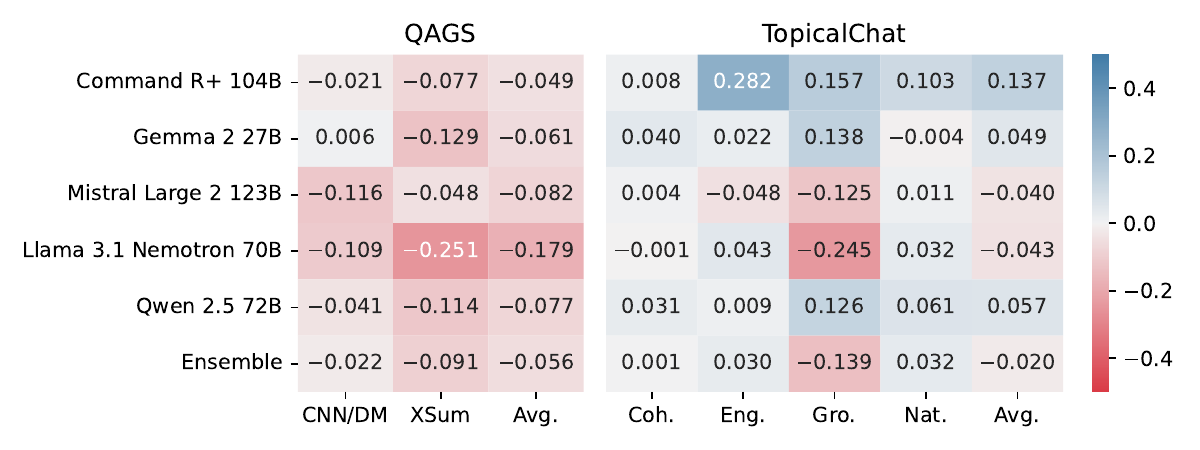}
    \caption{Results for Ablation 4 on QAGS and TopicalChat, where the rules section is removed from the prompt.}
    \label{fig:ablation4}
\end{figure*}

The detailed results of prompt ablation experiments are presented in Figure~\ref{fig:ablation1}--\ref{fig:ablation4}.
\subsection{Ensemble size ablations}
\label{app:ablations2}
\begin{figure*}[h!]
    \centering
    \includegraphics[width=\textwidth]{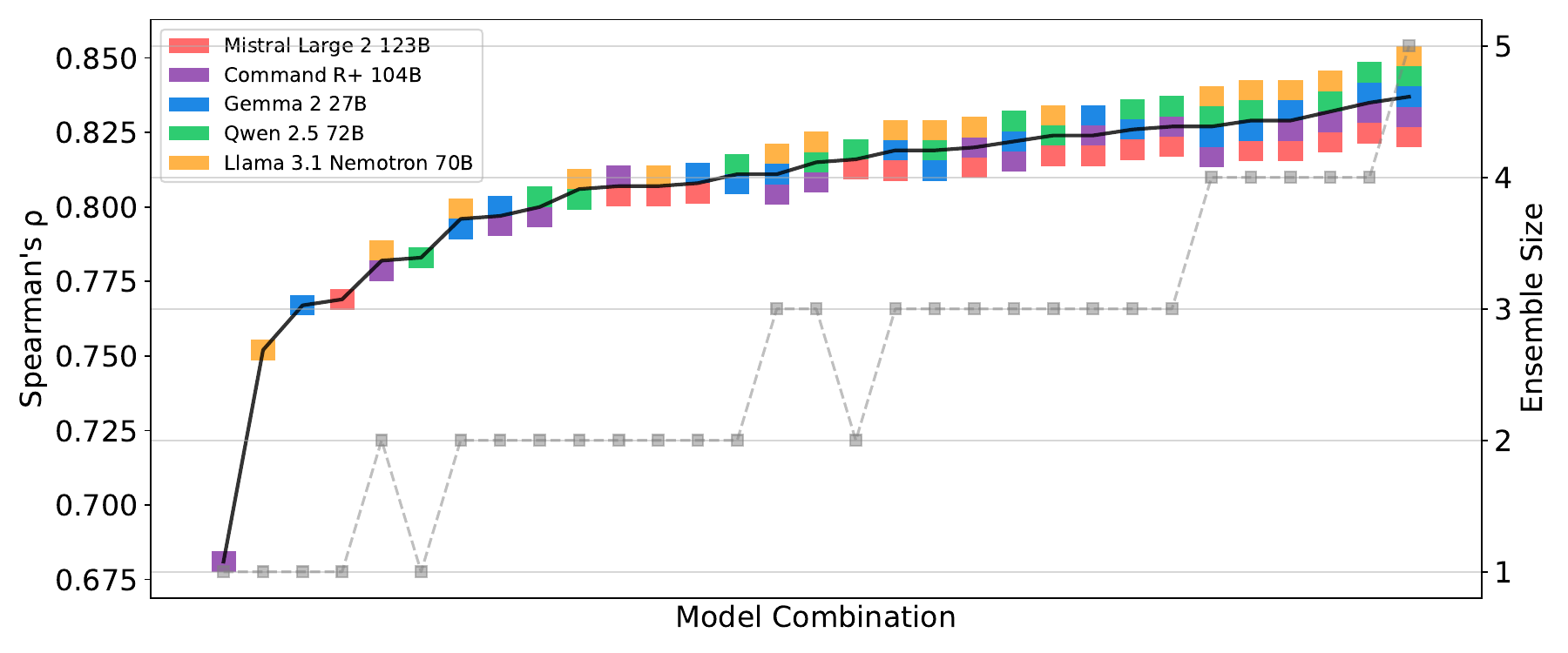}
    \caption{Effect of ensemble size on Spearman's $\rho$ correlations with human scores for the Wiki-DA dataset. Specific model combinations are represented by the colored patches.}
    \label{fig:ensemble_size_wikida}
\end{figure*}

\begin{figure*}[h!]
    \centering
    \includegraphics[width=\textwidth]{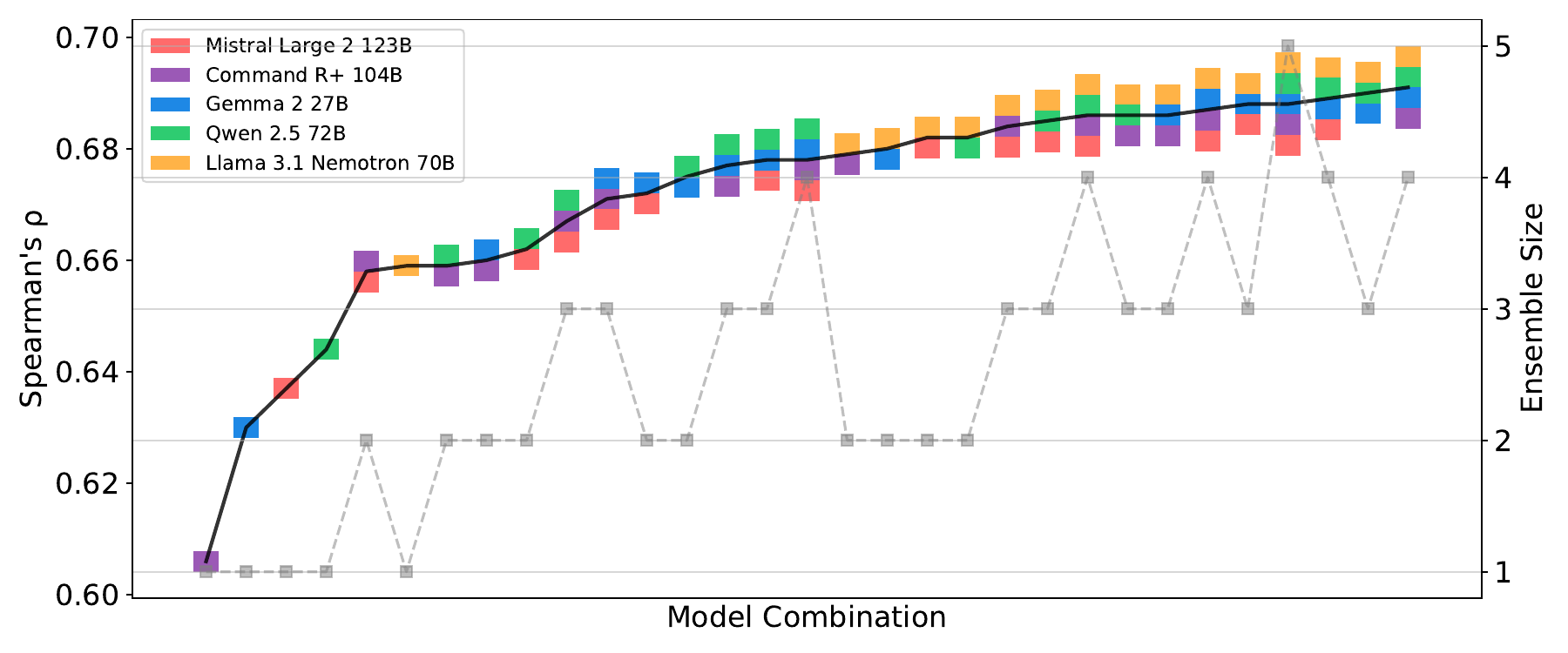}
    \caption{Effect of ensemble size on Spearman's $\rho$ correlations with human scores for the QAGS dataset.}
    \label{fig:ensemble_size_qags}
\end{figure*}

\begin{figure*}[h!]
    \centering
    \includegraphics[width=\textwidth]{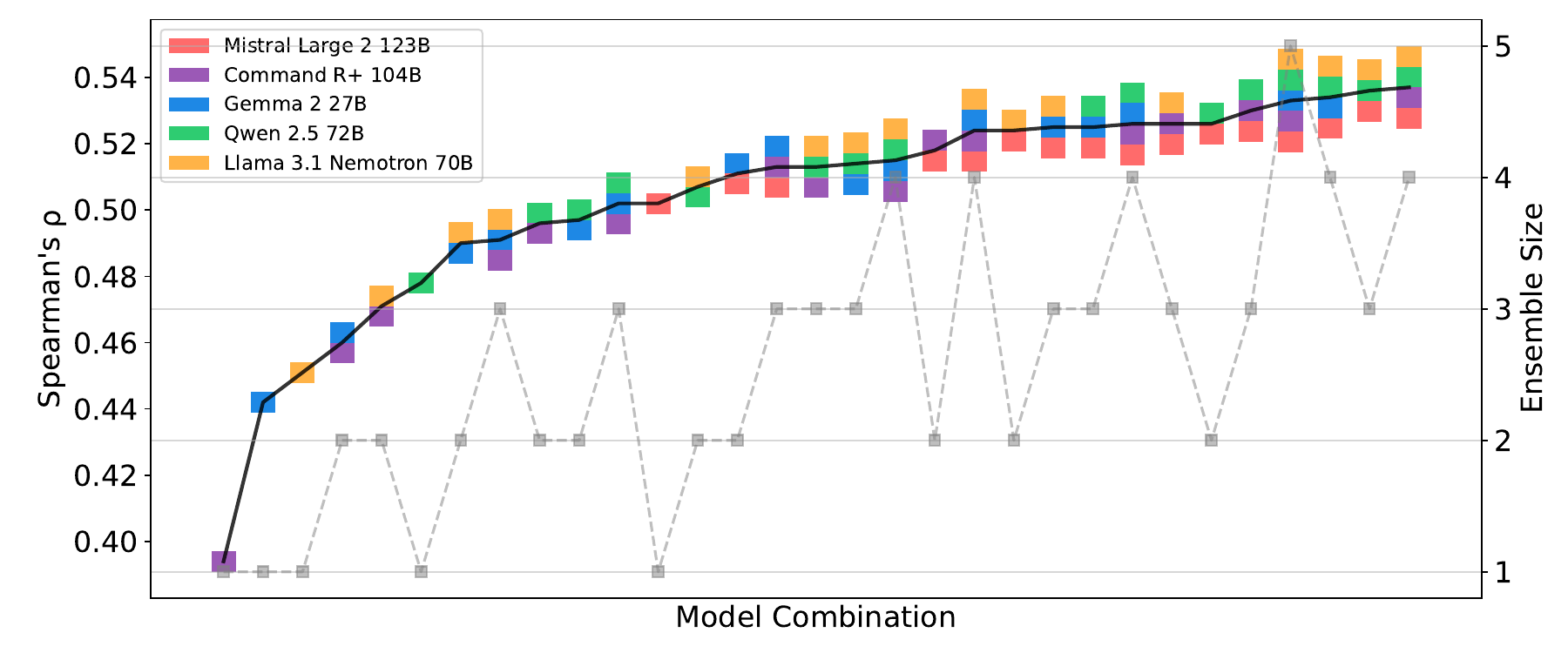}
    \caption{Effect of ensemble size on Spearman's $\rho$ correlations with human scores for the SummEval dataset.}
    \label{fig:ensemble_size_summeval}
\end{figure*}

\begin{figure*}[h!]
    \centering
    \includegraphics[width=\textwidth]{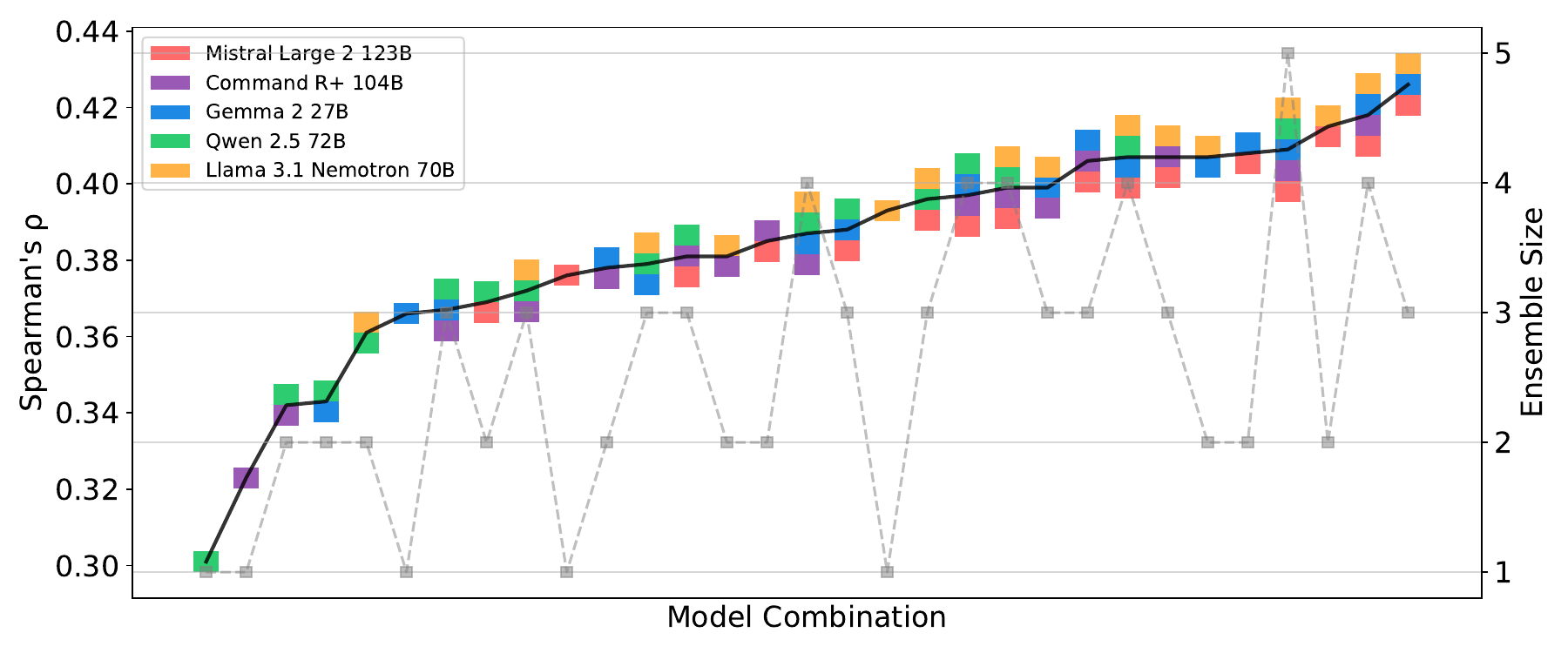}
    \caption{Effect of ensemble size on Spearman's $\rho$ correlations with human scores for the HANNA dataset.}
    \label{fig:ensemble_size_hanna}
\end{figure*}

\begin{figure*}[h!]
    \centering
    \includegraphics[width=\textwidth]{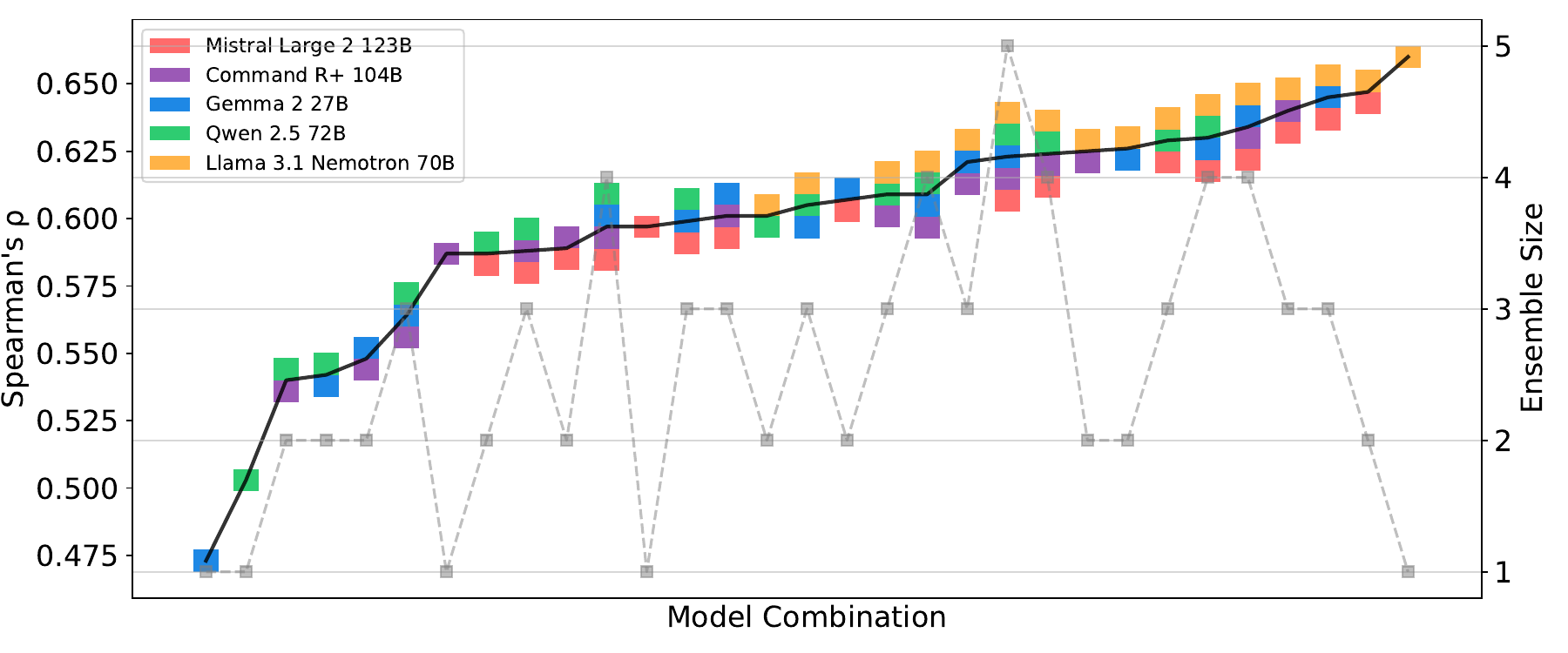}
    \caption{Effect of ensemble size on Spearman's $\rho$ correlations with human scores for the TopicalChat dataset.}
    \label{fig:ensemble_size_topical_chat}
\end{figure*}

\begin{figure*}[h!]
    \centering
    \includegraphics[width=\textwidth]{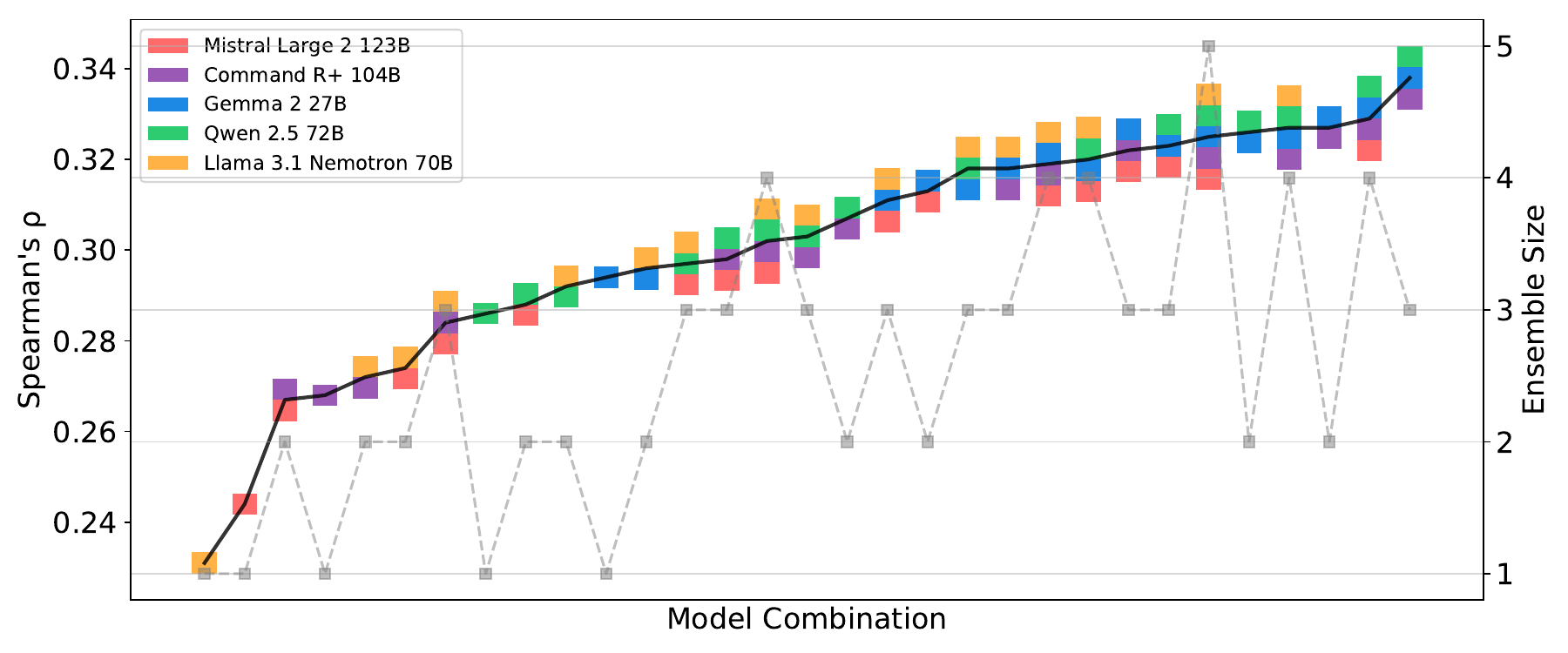}
    \caption{Effect of ensemble size on Spearman's $\rho$ correlations with human scores for the SFRES/SFHOT dataset.}
    \label{fig:ensemble_size_sfres}
\end{figure*}

The effect of the ensemble size on Spearman correlations are presented in Figures~\ref{fig:ensemble_size_wikida}--\ref{fig:ensemble_size_sfres}.

\section{Inter-annotator agreement between ensemble models}
\label{appx:llm_iaa}

To obtain additional insights into the variance between individual ensemble LLMs in their predicted overall scores, we compute pairwise inter-annotator agreements between the models on all meta-evaluation datasets.

\begin{figure*}[h!]
    \centering
    \includegraphics[width=\textwidth]{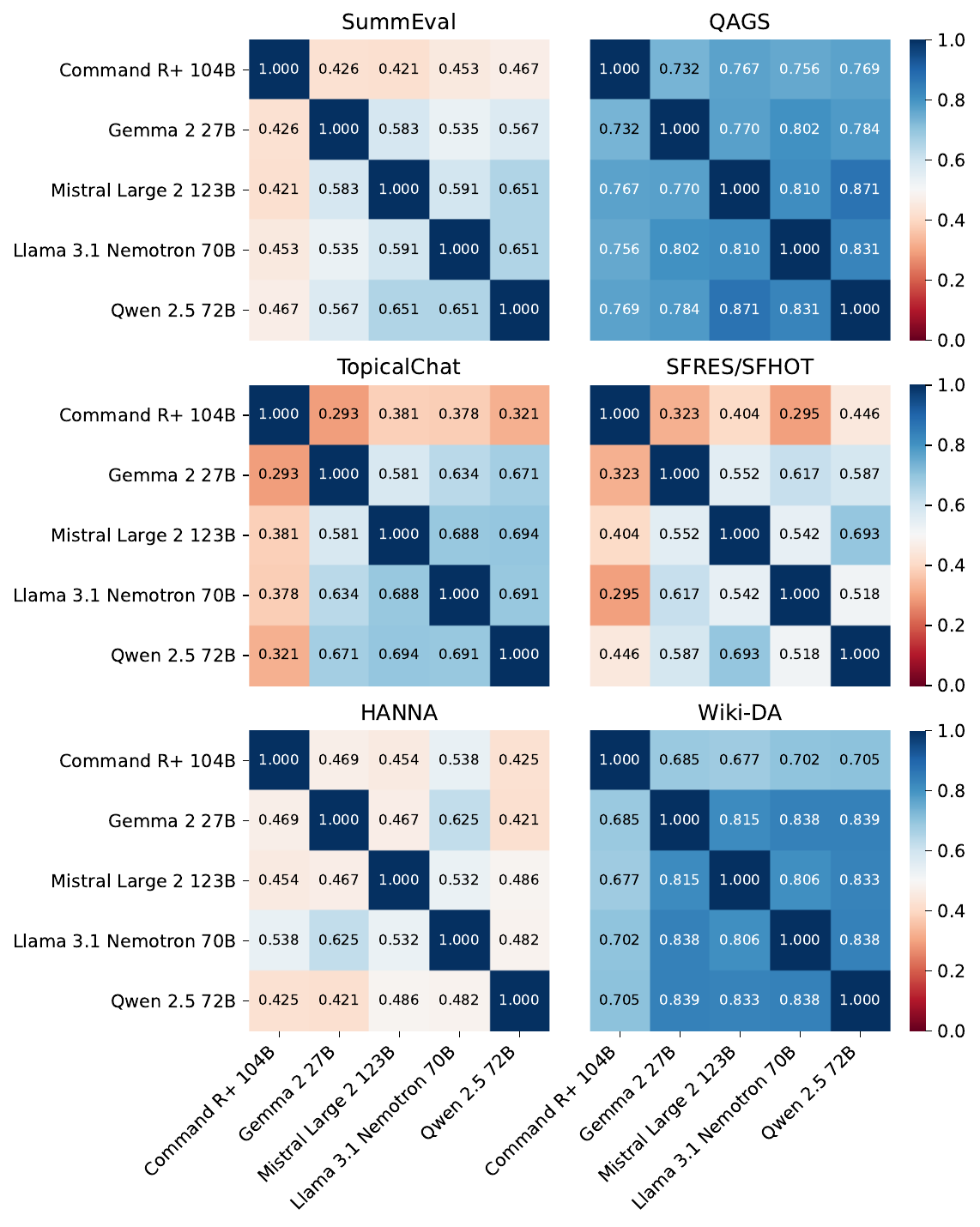}
    \caption{Pairwise Spearman ($\rho$) correlations of individual LLM scores for all meta-evaluation datasets. For each dataset, the correlations are computed over all evaluation aspects.}
    \label{fig:ensemble_spearman}
\end{figure*}

Figure \ref{fig:ensemble_spearman} shows the Spearman correlations for each LLM pair and dataset used in our meta-evaluation, where the correlations are calculated over all evaluation aspects in a given dataset. The agreement is generally high on the QAGS and Wiki-DA datasets, although substantially lower on other datasets. Note that individual models also achieved relatively high correlations with humans on these two datasets, which indicates that they might generally consist of examples (and possibly evaluation aspects) for which it is easier for LLMs to make decisions on their quality. In contrast, the models show relatively low agreement on HANNA, which could be explained by the more subjective nature of the story generation task and the corresponding evaluation aspects.\footnote{While such hypotheses could in principle be tested by comparing our results with the agreement between human annotators on a particular task, there are several issues that limit the reliability of such comparisons -- for each meta-evaluation dataset we use, at least two of the following hold: (1) the number of human annotators is different from the number of LLMs in our ensemble, (2) either the sets of human annotators differ between evaluated outputs, or it is not made explicit in the dataset (or the corresponding paper) whether this is the case, (3) rating scales are different from ours (often three levels, or even binary).} Across all tasks, Command R+ 104B tends to disagree more with other LLMs when compared to other pairwise agreements.

\begin{figure*}[h!]
    \centering
    \includegraphics[width=\textwidth]{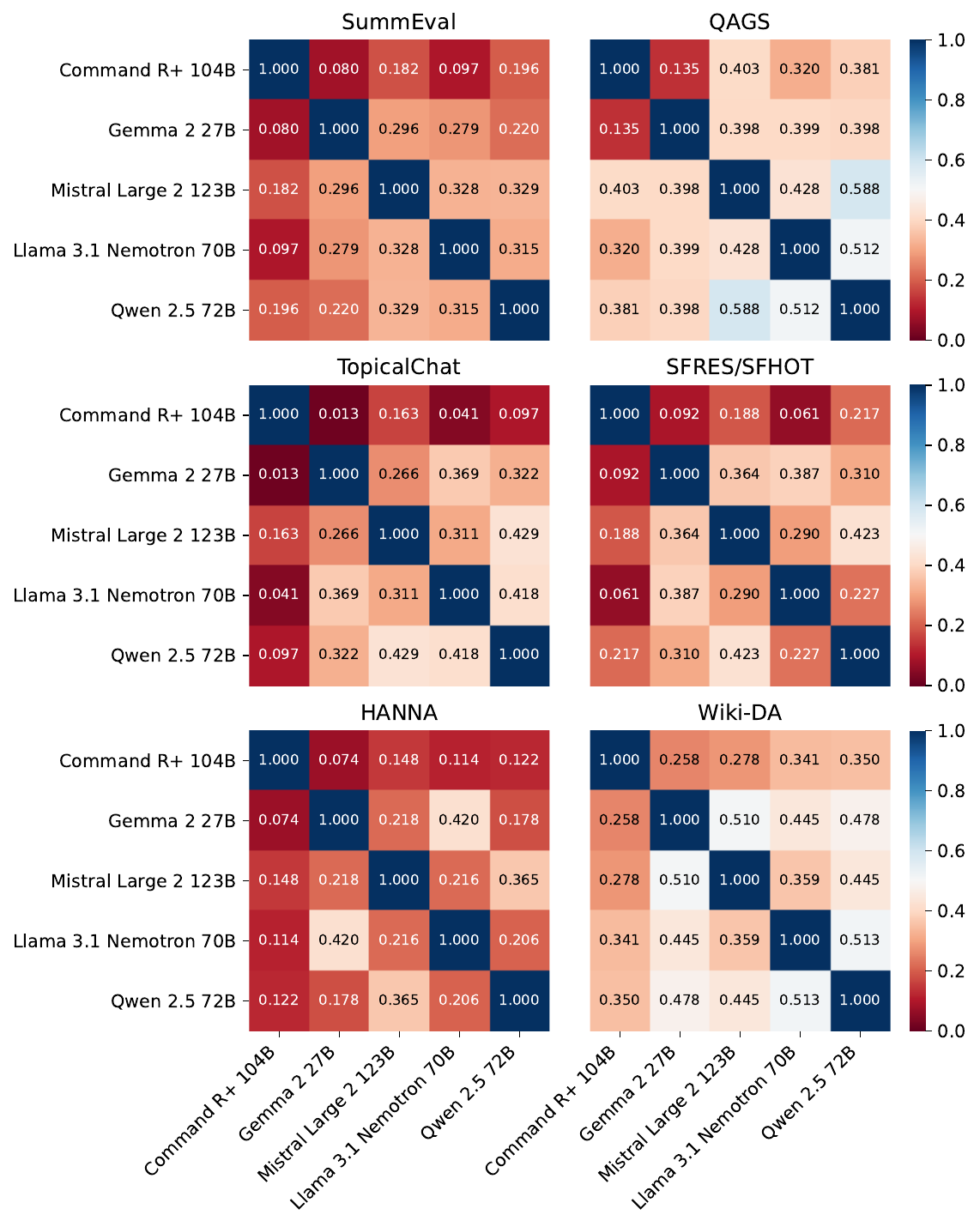}
    \caption{Pairwise inter-annotator agreements (Cohen's $\kappa$) of individual LLM scores for all meta-evaluation datasets. For each dataset, the coefficients are computed over all evaluation aspects.}
    \label{fig:ensemble_kappa}
\end{figure*}

To assess the agreement of the models on exact overall score predictions, we also measure pairwise Cohen's $\kappa$ (Figure \ref{fig:ensemble_kappa}), which treats the overall scores as categorical and therefore serves as a stricter measure. As expected, the agreements are lower compared to those measured by Spearman correlations, while the general trend is the same: the agreement on exact scores is higher on QAGS and Wiki-DA than on other datasets, while Command R+ 104B generally disagrees more with other LLMs across tasks.

Finally, we compute Krippendorff's $\alpha$ \citep{krippendorff2011computing}, which allows us to measure the agreement between all ensemble models, while also being applicable to ordinal data. In general, the agreements are low to moderate. Similarly to the pairwise agreements, we observe substantially higher agreements on QAGS and Wiki-DA, while the models agree the least on scoring generated stories in the HANNA dataset.

Overall, our analysis indicates that the predictions of individual models are sufficiently diverse to benefit from the combination in an ensemble without much redundancy.

\section{Score aggregation methods}
\label{appx:aggregation_methods}

Table \ref{tab:aggregation_methods} compares results of different approaches to aggregation of ensemble scores:
\begin{itemize}
    \item \textbf{Average} computes the final score for the evaluated output as a simple average of scores from all individual LLMs.
    \item \textbf{Average w/o outliers} first removes the outliers before computing the average. The score is considered an outlier if it differs from the average of other scores for the same example by at least two standard deviations and this difference is at least 1.
    \item \textbf{Median} computes the final score as a median of the scores from individual LLMs to disregard the effect of extreme scores.
    \item \textbf{Majority voting} selects the most frequent score from the individual LLMs for the given output.
    \item \textbf{Min} selects the minimum of individual scores as the final score. This corresponds to the most strict evaluation, i.e. typically the evaluation that detected the most errors or assigned highest severity levels to the identified errors.
\end{itemize}

\section{False positives analysis}
\label{app:false_positives}

To estimate the extent of potential over-annotation by \textsc{OpeNLGauge$_{ens}$} and its components, we analyze overall score predictions for output-aspect pairs $(y, a)$ which obtained maximum scores by all human annotators in a given dataset. We denote these examples $Y_{max}$ and assume that if $(y, a) \in Y_{max}$, then $y$ does not contain any errors with respect to aspect $a$.

Our analysis includes only those datasets and aspects that contain sufficiently fine-grained annotation scales (at least three levels), as it is unclear whether maximum scores on a binary rating scale reliably indicate a perceived lack of errors by human annotators. Additionally, we exclude datasets where the size of $Y_{max}$ is too small or even zero\footnote{For example, Wiki-DA contains only already aggregated scores on a scale between 0 and 100, with none of them equal to the maximum possible score. Although we could allow some deviation from the maximum score (e.g., maximum 5 points) to select $Y_{max}$, any such threshold would be arbitrary. Therefore, we exclude this dataset from the analysis.}. Given an output-aspect pair $(y, a)$$ \in Y_{max}$, we consider an LLM evaluation of $y$ with respect to $a$ an over-annotation if it contains one or more detected errors.

\begin{figure*}
    \centering
    \includegraphics[width=0.9\textwidth]{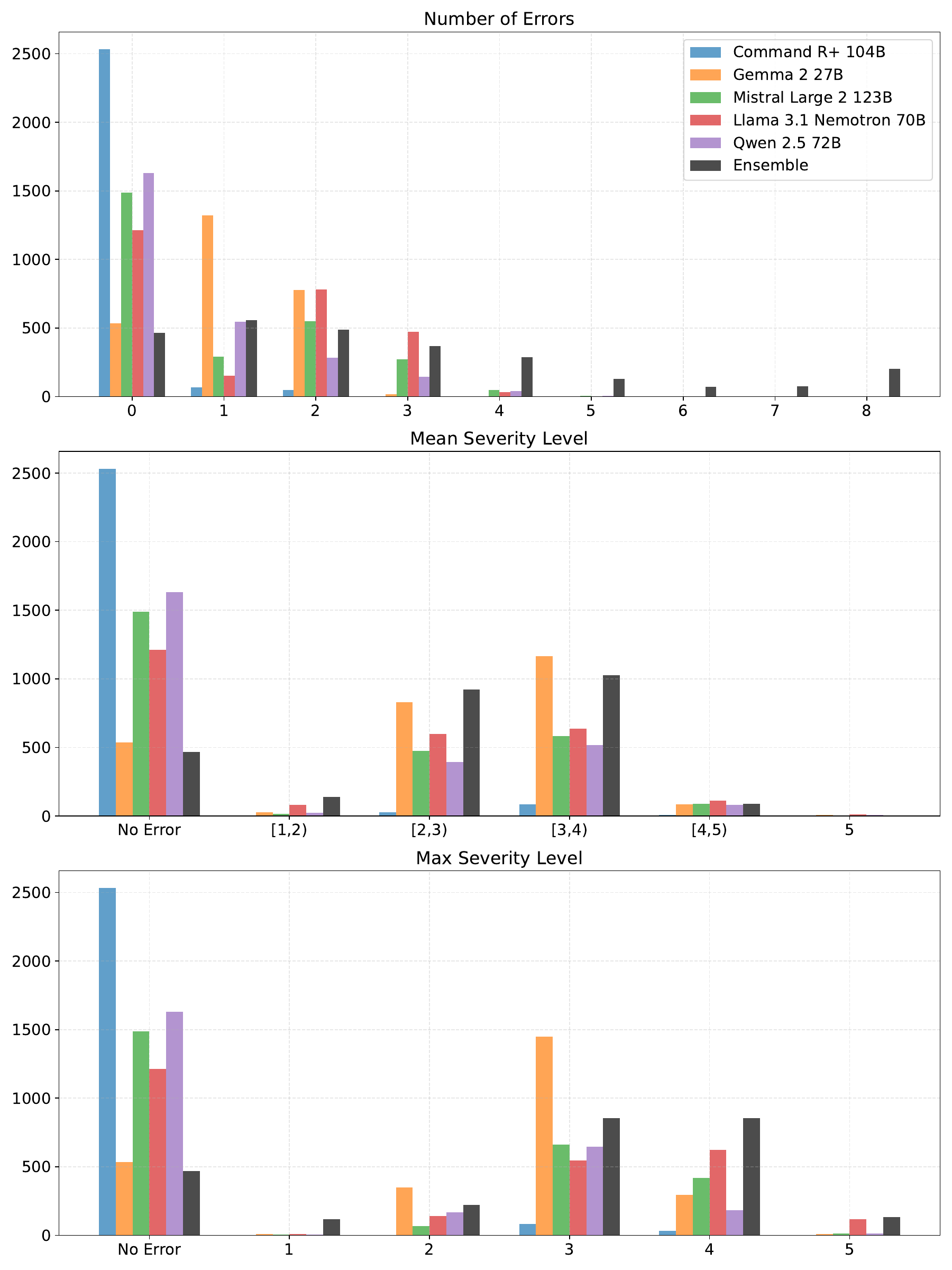}
    \caption{Distribution of errors detected by the ensemble LLMs in outputs rated with maximum score by human annotators in SummEval. \textbf{Top:} Frequencies of numbers of detected errors per evaluated output. \textbf{Middle:} Frequencies of mean severity levels assigned to detected error per output. Values larger than 0 are binned to ranges $[a, b)$, where $0 < a <= 5$ and $b = a + 1$. \textbf{Bottom:} Frequencies of maximum severity levels assigned to detected errors per output.}
    \label{fig:error_stats_summeval}
\end{figure*}

Figure \ref{fig:error_stats_summeval} shows the distribution of error counts (top), mean severity levels (middle) and maximum severity levels (bottom) per output, as predicted by \textsc{OpeNLGauge$_{ens}$} and its components for the $Y_{max}$ subset of SummEval. Most individual LLMs assess the outputs in $Y_{max}$ as error-free, particularly Command R+ 104B, which agrees almost perfectly with human annotators in this subset. Note that in general, Command R+ 104B tends to disagree the most with other LLMs, as discussed in Appendix \ref{appx:llm_iaa}, while also achieving the highest correlations with humans in evaluating factual consistency (Table \ref{tab:meta_eval_summeval}), which represents approximately half of the examples in $Y_{max}$. In contrast, the ensemble shows a tendency to include at least one error in most of its annotations for $Y_{max}$. This could be attributed to the merging procedure by the consolidator model, which aggregates errors from five different models and could lead to accumulation of errors. As expected, an increasing number of errors detected by a model is also reflected in the mean and maximum severity levels.

\begin{figure*}
    \centering
    \includegraphics[width=0.9\textwidth]{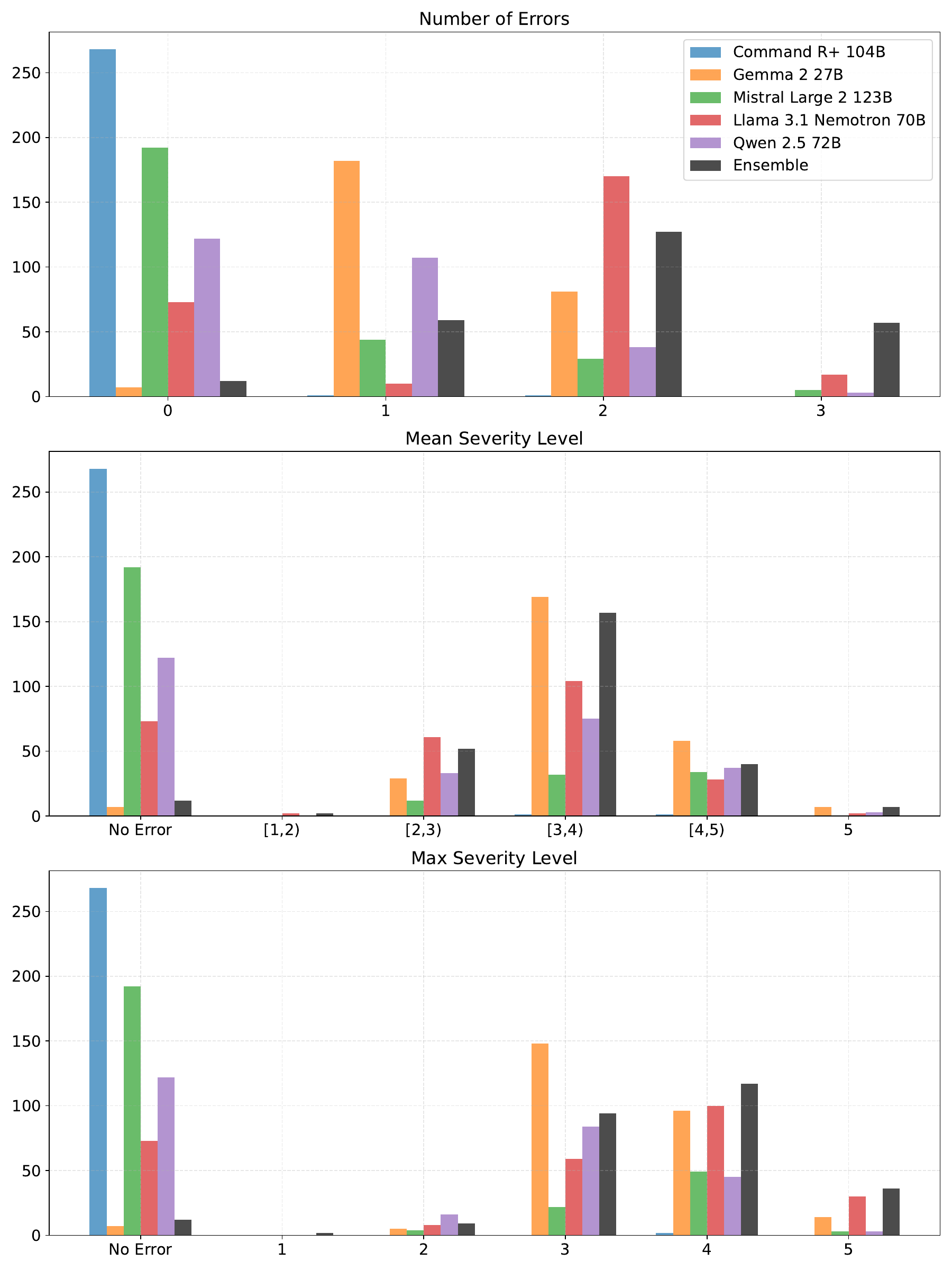}
    \caption{Distribution of errors detected by the ensemble LLMs in outputs rated with maximum score by human annotators in TopicalChat. Note that groundedness evaluations are excluded from the analysis, as the dataset contains only binary ratings for this aspect. \textbf{Top:} Frequencies of numbers of detected errors per evaluated output. \textbf{Middle:} Frequencies of mean severity levels assigned to detected error per output. Values larger than 0 are binned to ranges $[a, b)$, where $0 < a <= 5$ and $b = a + 1$. \textbf{Bottom:} Frequencies of maximum severity levels assigned to detected errors per output.}
    \label{fig:error_stats_topical_chat}
\end{figure*}

The error annotations for TopicalChat (Figure \ref{fig:error_stats_topical_chat}) show a similar pattern, although a larger proportion of individual models have a tendency to detect one or more errors in this case.

\section{Output examples}
\label{appx:output_examples}

Figures~\ref{fig:exp2}--\ref{fig:exp4} show additional output examples for RDF-to-text, dialogue response generation, and summarization tasks. 

\newpage
\begin{figure*}[htbp]
    \newpage
    \begin{lstlisting}[language=,caption={Annotator prompt template for the data-to-text task.},captionpos=b,label={lst:prompt-annotator}]
    ### Instructions
    Your task is to evaluate an output of data-to-text task, where the model was instructed to write a single-paragraph description of a venue based on the given data. The data consist of a set of attribute-value pairs in the form 'attribute: value'.
    Based on the given data and the generated text, identify errors in the text with respect to {{ aspect_name }} (described below).
    For each error, determine its severity on a scale from 1 to 5, where 1 is the least severe and 5 is the most severe.
    
    Definition of {{ aspect_name }}:
    {{ aspect_definition }}
    
    Rules:
    Do not make assumptions and do not bring in external knowledge not present in the provided context.
    Identify only the errors related to the {{ aspect_name }} of the text. Do not consider other aspects like {{ negative_aspects }}!
    If there are no errors related to {{ aspect_name }} in the text, you should output 'No Error' and provide 'Excellent' score.
    
    Steps:
    1. Carefully read the data and identify the main attributes and their values.
    2. Read the generated text and compare it with the source data with respect to {{ aspect_name }}.
    3. If the text contains any error that negatively affects its {{ aspect_name }}, identify its exact location (specific word or phrase), explain why it is considered an error, and determine the severity of the error.
    4. Finally, provide an overall score for the {{ aspect_name }} of the text. The score should be a label on the following scale (lowest to highest): 'Unacceptable', 'Poor', 'Fair', 'Good', 'Excellent'. The score 'Unacceptable' indicates that the text is {{ min_score_desc }}, while 'Excellent' indicates that the text is {{ max_score_desc }}.
    
    ### Data
    {{ input }}
    
    ### Generated Text
    {{ output }}
    
    ### Output format:
    Generate your output exactly in this format:
    ```
    Error 1:
    Location: <location of the error - the exact word or phrase in the response>
    Explanation: <explanation for the error, including the reason why it is considered {{ aspect_name }} issue>
    Severity: <integer from 1 to 5>
    
    Error 2:
    ...
    
    Overall score: <one of: Unacceptable, Poor, Fair, Good, Excellent>
    Explanation of the score: <explanation of the score>
    ```
    \end{lstlisting}
\end{figure*}

\newpage
\begin{figure*}[htbp]
    \centering
    \begin{lstlisting}[language=,caption={Consolidator prompt template for merging of individual annotations.},captionpos=b,label={lst:prompt-supervisor}]
    ### Instructions
    You are given multiple error annotation sets for an AI model output. Your task is to merge the annotation sets to a single final annotation set.
    The result shouldn't contain any duplicates. If there are multiple error annotations for approximately the same location that describe the same issue, you should merge them into single location. Otherwise, the error annotations should be as granular as possible. If there are multiple different locations with the same issue, each should have its own error annotation. Likewise, if there are multiple issues with respect to the same location, each should have its own error annotation. Use the following guidelines:
    * Each error annotation should describe a single issue.
    * Merge only annotations where the locations have significant overlap.
    * When merging multiple locations, choose a single span from the output text that covers the locations from merged annotations.
    * Never include multiple spans from the annotations under the same "Location" line.
    * Do not include any other text in "Location" line than the text that is actually in the output, except for annotations that mention omissions or similar issues.
    * When merging explanations, combine the most relevant information from the merged annotations.
    * Severity levels range from 1 to 5 from least severe to most severe. Use the most severe level from the merged annotations.
    * Final annotation set should not include more than 8 error annotations. If there are more than that, use only the most severe ones.
    * Make the final annotation set as concise as possible in terms of number of error annotations.
    
    Don't use any markdown formatting. Generate merged error annotations in this format, without any additional text:
    ```
    Error 1:
    Location: <span of text from the output, or None if not applicable>
    Explanation: <explanation>
    Severity: <severity level>
    ```
    
    ### Model output
    {{ output }}
    
    ### Error annotations
    {{ annotations }}
    \end{lstlisting}
\end{figure*}

\newpage
\begin{figure*}[htbp]
    \newpage
    \begin{lstlisting}[language=,caption={Prompt template for the fine-tuned \textsc{OpeNLGauge$_{ft}$} metric.},captionpos=b,label={lst:prompt-train}]
    ### Task
    Your task is to evaluate a model output for {{ task_name }} task with respect to {{ aspect_name }}. {{ extra_task_info }}
    
    ### Aspect Definition
    {{ aspect_name }} - {{ aspect_definition }}
    
    ### Dialogue history
    {{ input }}
    {% if context %}
    ### Knowledge
    {{ context }}
    {% endif %}
    ### Response
    {{ output }}
    
    ### Instructions
    For any error in the output, identify its location, assign a severity level and provide an explanation. Report at most 8 errors. If there are more errors, report only the most severe ones. Finally, provide an overall score between 0 and 100 for {{ aspect_name }} of the output.
    \end{lstlisting}
\end{figure*}

\begin{table*}[htbp]
\small \centering
\begin{tabular}{lcccccccccccc}
\toprule
\multirow{2}{*}{\textbf{Metric}} & \multicolumn{2}{c}{\textbf{Consistency}} & \multicolumn{2}{c}{\textbf{Coherence}} & \multicolumn{2}{c}{\textbf{Relevance}} & \multicolumn{2}{c}{\textbf{Fluency}} & \multicolumn{2}{c}{\textbf{Average}} \\
 & $\rho$ & $\tau$ & $\rho$ & $\tau$ & $\rho$ & $\tau$ & $\rho$ & $\tau$ & $\rho$ & $\tau$ \\
\midrule
ROUGE-1 & 0.167 & 0.126 & 0.160 & 0.130 & 0.115 & 0.094 & 0.326 & 0.252 & 0.192 & 0.150 \\
ROUGE-2 & 0.184 & 0.139 & 0.187 & 0.155 & 0.159 & 0.128 & 0.290 & 0.219 & 0.205 & 0.161 \\
ROUGE-L & 0.128 & 0.099 & 0.115 & 0.092 & 0.105 & 0.084 & 0.311 & 0.237 & 0.165 & 0.128 \\
BERTScore & 0.151 & 0.122 & 0.285 & 0.220 & 0.302 & 0.232 & 0.186 & 0.154 & 0.231 & 0.182 \\
MOVERSscore & 0.159 & 0.118 & 0.157 & 0.127 & 0.129 & 0.105 & 0.318 & 0.244 & 0.191 & 0.148 \\
\midrule
BARTScore & 0.266 & 0.220 & 0.474 & 0.367 & 0.318 & 0.243 & 0.258 & 0.214 & 0.329 & 0.261 \\
UniEval & 0.446 & 0.371 & 0.575 & 0.442 & 0.426 & 0.325 & 0.449 & 0.371 & 0.474 & 0.377 \\
\midrule
G-Eval (GPT-3.5) & 0.386 & 0.318 & 0.440 & 0.335 & 0.385 & 0.293 & 0.424 & 0.347 & 0.409 & 0.323 \\
G-Eval (GPT-4) & 0.507 & 0.425 & \underline{0.582} & 0.457 & \textbf{0.548} & \textbf{0.433} & 0.455 & 0.378 & 0.523 & 0.423 \\
LLM Evaluation (GPT-3.5) & 0.393 & 0.331 & 0.459 & 0.371 & 0.455 & 0.363 & 0.355 & 0.296 & 0.415 & 0.340 \\
LLM Evaluation (GPT-4) & 0.531 & 0.464 & 0.540 & 0.434 & 0.491 & 0.395 & \underline{0.480} & \underline{0.409} & 0.511 & 0.426 \\
\midrule
X-Eval & 0.428 & 0.340 & 0.530 & 0.382 & 0.500 & 0.361 & 0.461 & 0.365 & 0.480 & 0.362 \\
Prometheus & 0.150 & 0.137 & 0.150 & 0.126 & 0.164 & 0.138 & 0.189 & 0.168 & 0.163 & 0.142 \\
AUTO-J & 0.131 & 0.121 & 0.245 & 0.203 & 0.262 & 0.222 & 0.154 & 0.141 & 0.198 & 0.172 \\
TIGERScore & 0.427 & 0.387 & 0.381 & 0.318 & 0.366 & 0.304 & 0.363 & 0.327 & 0.384 & 0.334 \\
InstructScore & 0.232 & 0.213 & 0.328 & 0.276 & 0.211 & 0.179 & 0.260 & 0.237 & 0.258 & 0.226 \\
Themis & 0.600 & 0.566 & 0.566 & \textbf{0.485} & 0.474 & \underline{0.412} & \textbf{0.571} & \textbf{0.533} & \textbf{0.553} & \textbf{0.499} \\
\midrule
\textsc{OpeNLGauge$_{ens}$} & 0.548 & 0.470 & \textbf{0.604} & \underline{0.462} & 0.513 & 0.389 & 0.470 & 0.389 & \underline{0.534} & 0.427 \\
 \textbullet \space Command R+ 104B & \textbf{0.633} & \textbf{0.603} & 0.239 & 0.203 & 0.360 & 0.302 & 0.347 & 0.320 & 0.395 & 0.357 \\
 \textbullet \space Gemma 2 27B & 0.459 & 0.421 & 0.455 & 0.386 & 0.428 & 0.358 & 0.427 & 0.386 & 0.442 & 0.388 \\
 \textbullet \space Llama 3.1 Nemotron 70B & 0.559 & 0.517 & 0.469 & 0.391 & 0.419 & 0.356 & 0.358 & 0.325 & 0.451 & 0.397 \\
 \textbullet \space Mistral Large 2 123B & \underline{0.627} & \underline{0.590} & 0.528 & 0.434 & 0.456 & 0.375 & 0.398 & 0.359 & 0.502 & \underline{0.439} \\
 \textbullet \space Qwen 2.5 72B & 0.567 & 0.521 & 0.525 & 0.433 & 0.388 & 0.317 & 0.433 & 0.389 & 0.478 & 0.415 \\
\midrule
Llama 3.1 8B & 0.181 & 0.165 & 0.176 & 0.150 & 0.156 & 0.127 & 0.232 & 0.218 & 0.186 & 0.165 \\
\textsc{OpeNLGauge$_{ft}$} & 0.527 & 0.453 & 0.561 & 0.441 & \underline{0.514} & 0.408 & 0.407 & 0.349 & 0.502 & 0.413 \\
\bottomrule
\end{tabular}
\caption{Segment-level Spearman ($\rho$) and Kendall ($\tau$) correlations of different metrics on SummEval.}
\label{tab:meta_eval_summeval}
\end{table*}

\begin{table*}[htbp]
\small
\centering
\resizebox{\textwidth}{!}{%
\begin{tabular}{lcccccccccc}
\toprule
\multirow{2}{*}{\textbf{Metric}} & \multicolumn{2}{c}{\textbf{Groundedness}} & \multicolumn{2}{c}{\textbf{Coherence}} & \multicolumn{2}{c}{\textbf{Engagingness}} & \multicolumn{2}{c}{\textbf{Naturalness}} & \multicolumn{2}{c}{\textbf{Average}} \\
 & $r$ & $\rho$ & $r$ & $\rho$ & $r$ & $\rho$ & $r$ & $\rho$ & $r$ & $\rho$ \\
\midrule
ROUGE-L & 0.193 & 0.203 & 0.176 & 0.146 & 0.295 & 0.300 & 0.310 & 0.327 & 0.243 & 0.244 \\
BLEU-4 & 0.131 & 0.235 & 0.180 & 0.175 & 0.232 & 0.316 & 0.213 & 0.310 & 0.189 & 0.259 \\
METEOR & 0.250 & 0.302 & 0.212 & 0.191 & 0.367 & 0.439 & 0.333 & 0.391 & 0.290 & 0.331 \\
BERTScore & 0.214 & 0.233 & 0.226 & 0.209 & 0.317 & 0.335 & 0.291 & 0.317 & 0.262 & 0.273 \\
\midrule
USR & 0.416 & 0.377 & 0.337 & 0.325 & 0.456 & 0.465 & 0.222 & 0.447 & 0.358 & 0.403 \\x
UniEval & 0.602 & 0.455 & 0.455 & 0.330 & 0.573 & 0.430 & 0.577 & 0.453 & 0.552 & 0.417 \\
\midrule
G-Eval (GPT-3.5) & 0.586 & 0.567 & 0.519 & 0.544 & 0.660 & 0.691 & 0.532 & 0.539 & 0.574 & 0.585 \\
G-Eval (GPT-4) & 0.531 & 0.551 & 0.594 & 0.605 & 0.627 & 0.631 & 0.549 & 0.565 & 0.575 & 0.588 \\
LLM Evaluation (GPT-3.5) & 0.653 & 0.581 & 0.550 & 0.531 & 0.651 & 0.648 & 0.515 & 0.550 & 0.592 & 0.578 \\
LLM Evaluation (GPT-4) & \textbf{0.810} & \underline{0.786} & \textbf{0.680} & \textbf{0.680} & \textbf{0.822} & \textbf{0.779} & \textbf{0.769} & \textbf{0.739} & \textbf{0.770} & \textbf{0.746} \\
\midrule
X-Eval & 0.734 & 0.728 & 0.558 & 0.622 & 0.449 & 0.593 & 0.417 & 0.478 & 0.539 & 0.605 \\
Prometheus & 0.437 & 0.412 & 0.451 & 0.465 & 0.495 & 0.473 & 0.355 & 0.384 & 0.435 & 0.434 \\
AUTO-J & 0.339 & 0.357 & 0.452 & 0.449 & 0.490 & 0.459 & 0.425 & 0.437 & 0.427 & 0.425 \\
TIGERScore & 0.137 & 0.138 & 0.417 & 0.438 & 0.328 & 0.333 & 0.455 & 0.477 & 0.334 & 0.346 \\
InstructScore & 0.140 & 0.102 & 0.299 & 0.297 & 0.264 & 0.233 & 0.374 & 0.332 & 0.269 & 0.241 \\
Themis & 0.778 & 0.761 & \underline{0.639} & \underline{0.644} & \underline{0.790} & \underline{0.766} & \underline{0.727} & \underline{0.729} & \underline{0.733} & \underline{0.725} \\
\midrule
\textsc{OpeNLGauge$_{ens}$} & 0.704 & 0.697 & 0.622 & 0.621 & 0.675 & 0.692 & 0.599 & 0.604 & 0.649 & 0.653 \\
 \textbullet \space Command R+ 104B & 0.383 & 0.368 & 0.463 & 0.453 & 0.262 & 0.259 & 0.421 & 0.398 & 0.386 & 0.374 \\
 \textbullet \space Gemma 2 27B & 0.332 & 0.366 & 0.465 & 0.481 & 0.549 & 0.562 & 0.489 & 0.515 & 0.459 & 0.484 \\
 \textbullet \space Llama 3.1 Nemotron 70B & \underline{0.781} & \textbf{0.791} & 0.600 & 0.630 & 0.655 & 0.686 & 0.506 & 0.532 & 0.621 & 0.645 \\
 \textbullet \space Mistral Large 2 123B & 0.658 & 0.648 & 0.541 & 0.554 & 0.645 & 0.659 & 0.509 & 0.529 & 0.586 & 0.596 \\
 \textbullet \space Qwen 2.5 72B & 0.467 & 0.460 & 0.480 & 0.521 & 0.500 & 0.516 & 0.468 & 0.488 & 0.496 & 0.514 \\
\midrule
Llama 3.1 8B & 0.374 & 0.362 & 0.225 & 0.228 & 0.415 & 0.408 & 0.222 & 0.238 & 0.309 & 0.309 \\
\textsc{OpeNLGauge$_{ft}$} & 0.485 & 0.538 & 0.531 & 0.575 & 0.622 & 0.650 & 0.522 & 0.547 & 0.540 & 0.578 \\
\bottomrule
\end{tabular}
}
\caption{Segment-level Pearson ($r$) and Spearman ($\rho$) correlations of different metrics on TopicalChat.}
\label{tab:meta_eval_topical_chat}
\end{table*}

\begin{table*}[htbp]
\small
\centering
\begin{tabular}{lccccc}
\toprule
\multirow{2}{*}{\textbf{Metric}} & \multicolumn{2}{c}{\textbf{SFRES}} & \multicolumn{2}{c}{\textbf{SFHOT}} & \multirow{2}{*}{\textbf{Average}} \\
 & \textbf{Inf.} & \textbf{Nat.} & \textbf{Inf.} & \textbf{Nat.} &  \\
\midrule
ROUGE-1 & 0.115 & 0.170 & 0.118 & 0.196 & 0.150 \\
ROUGE-L & 0.103 & 0.169 & 0.110 & 0.186 & 0.142 \\
BERTScore & 0.156 & 0.219 & 0.135 & 0.178 & 0.172 \\
MOVERScore & 0.153 & 0.190 & 0.172 & 0.242 & 0.189 \\
\midrule
BARTScore & 0.238 & 0.289 & 0.235 & 0.288 & 0.263 \\
UniEval & 0.225 & 0.333 & 0.249 & 0.320 & 0.282 \\
\midrule
GPTScore & 0.232 & 0.190 & 0.184 & 0.036 & 0.161 \\
G-Eval (GPT-4) & 0.189 & 0.351 & 0.198 & 0.338 & 0.269 \\
LLM Evaluation (GPT-3.5) & \underline{0.304} & 0.385 & 0.242 & 0.294 & 0.306 \\
LLM Evaluation (GPT-4) & 0.213 & \underline{0.405} & \textbf{0.302} & \underline{0.359} & \underline{0.320} \\
\midrule
Prometheus & 0.161 & 0.150 & 0.211 & 0.169 & 0.173 \\
Auto-J & 0.179 & 0.084 & 0.176 & 0.127 & 0.141 \\
TIGERScore & 0.160 & 0.221 & 0.215 & 0.204 & 0.200 \\
InstructScore & 0.194 & 0.300 & 0.222 & 0.273 & 0.247 \\
Themis & 0.298 & 0.395 & 0.259 & \textbf{0.380} & \textbf{0.333} \\
\midrule
\textsc{OpeNLGauge$_{ens}$} & 0.234 & \textbf{0.415} & 0.205 & 0.341 & 0.299 \\
 \textbullet \space Command R+ 104B & 0.239 & 0.298 & 0.215 & 0.311 & 0.266 \\
 \textbullet \space Gemma 2 27B & 0.254 & 0.334 & \underline{0.275} & 0.317 & 0.295 \\
 \textbullet \space Llama 3.1 Nemotron 70B & 0.178 & 0.284 & 0.047 & 0.209 & 0.179 \\
 \textbullet \space Mistral Large 2 123B & 0.129 & 0.359 & 0.226 & 0.281 & 0.249 \\
 \textbullet \space Qwen 2.5 72B & 0.226 & 0.347 & 0.245 & 0.315 & 0.283 \\
\midrule
Llama 3.1 8B & 0.003 & 0.081 & 0.071 & 0.094 & 0.108 \\
\textsc{OpeNLGauge$_{ft}$} & \textbf{0.354} & 0.354 & 0.238 & 0.315 & 0.315 \\
\bottomrule
\end{tabular}
\caption{Segment-level Spearman ($\rho$) correlations of different metrics on SFRES and SFHOT. \textbf{Inf.} = informativeness, \textbf{Nat.} = naturalness.}
\label{tab:meta_eval_sfres_sfhot}
\end{table*}

\begin{table*}\small
\centering
\begin{tabular}{lccccccc}
\toprule
\textbf{Metric} & \textbf{Coh.} & \textbf{Rel.} & \textbf{Eng.} & \textbf{Emp.} & \textbf{Sur.} & \textbf{Com.} & \textbf{Avg.} \\
\midrule
BLEU & 0.539 & 0.514 & 0.483 & 0.410 & 0.471 & 0.516 & 0.489 \\
ROUGE-1 & \textbf{0.567} & 0.518 & \underline{0.529} & \textbf{0.450} & \textbf{0.490} & \textbf{0.591} & \textbf{0.524} \\
METEOR & 0.560 & 0.522 & 0.510 & 0.435 & \underline{0.488} & 0.555 & 0.512 \\
MoverScore & 0.551 & 0.523 & 0.495 & 0.418 & 0.478 & 0.530 & 0.499 \\
BERTScore & \underline{0.566} & \underline{0.531} & 0.520 & 0.441 & \underline{0.488} & 0.563 & \underline{0.518} \\
BARTScore & 0.501 & 0.467 & 0.465 & 0.416 & 0.436 & 0.488 & 0.462 \\
\midrule
\textsc{OpeNLGauge$_{ens}$} & 0.528 & \textbf{0.559} & \textbf{0.538} & 0.434 & 0.343 & \textbf{0.591} & 0.499 \\
 \textbullet \space Command R+ 104B & 0.412 & 0.383 & 0.420 & 0.330 & 0.227 & 0.344 & 0.353 \\
 \textbullet \space Gemma 2 27B & 0.453 & 0.445 & 0.461 & 0.356 & 0.323 & 0.521 & 0.427 \\
 \textbullet \space Llama 3.1 Nemotron 70B & 0.500 & 0.508 & 0.497 & 0.380 & 0.332 & \underline{0.565} & 0.464 \\
 \textbullet \space Mistral Large 2 123B & 0.372 & 0.490 & 0.425 & 0.373 & 0.334 & 0.507 & 0.417 \\
 \textbullet \space Qwen 2.5 72B & 0.407 & 0.484 & 0.427 & 0.277 & -0.012 & 0.507 & 0.348 \\
\midrule
Llama 3.1 8B & 0.119 & 0.344 & 0.154 & 0.094 & 0.080 & 0.212 & 0.167 \\
\textsc{OpeNLGauge$_{ft}$} & 0.448 & 0.523 & 0.517 & \underline{0.444} & 0.404 & 0.555 & 0.482\\
\bottomrule
\end{tabular}
\caption{Segment-level Pearson (\(r\)) correlations of different metrics on HANNA. \textbf{Coh.} = coherence, \textbf{Rel.} = relevance, \textbf{Eng.} = engagement, \textbf{Emp.} = empathy, \textbf{Sur.} = surprise, \textbf{Com.} = complexity, \textbf{Avg.} = average.}
\label{tab:meta_eval_hanna_r}
\end{table*}

\begin{table*}
\centering \small
\begin{tabular}{lccccccc}
\toprule
\textbf{Metric} & \textbf{Coh.} & \textbf{Rel.} & \textbf{Eng.} & \textbf{Emp.} & \textbf{Sur.} & \textbf{Com.} & \textbf{Avg.} \\
\midrule
BLEU & 0.339 & 0.292 & 0.356 & 0.315 & 0.299 & 0.414 & 0.336 \\
ROUGE-1 & 0.389 & 0.330 & 0.416 & 0.354 & \textbf{0.355} & \underline{0.503} & 0.391 \\
METEOR & 0.378 & 0.310 & 0.412 & 0.366 & \underline{0.354} & \textbf{0.505} & 0.387 \\
MoverScore & 0.392 & 0.385 & 0.420 & 0.331 & 0.321 & 0.473 & 0.387 \\
BERTScore & 0.372 & 0.355 & 0.415 & 0.356 & 0.320 & 0.469 & 0.381 \\
BARTScore & 0.259 & 0.249 & 0.291 & 0.287 & 0.227 & 0.294 & 0.268 \\
\midrule
\textsc{OpeNLGauge$_{ens}$} & 0.393 & \underline{0.474} & \underline{0.452} & \underline{0.367} & 0.276 & 0.489 & \underline{0.409} \\
 \textbullet \space Command R+ 104B & 0.325 & 0.362 & 0.372 & 0.320 & 0.215 & 0.348 & 0.324 \\
 \textbullet \space Gemma 2 27B & 0.381 & 0.383 & 0.400 & 0.310 & 0.278 & 0.445 & 0.366 \\
 \textbullet \space Llama 3.1 Nemotron 70B & \textbf{0.429} & 0.445 & 0.427 & 0.328 & 0.263 & 0.464 & 0.393 \\
 \textbullet \space Mistral Large 2 123B & 0.294 & 0.415 & 0.389 & 0.349 & 0.337 & 0.474 & 0.376 \\
 \textbullet \space Qwen 2.5 72B & 0.344 & 0.423 & 0.364 & 0.262 & -0.006 & 0.416 & 0.301 \\
\midrule
Llama 3.1 8B & 0.093 & 0.334 & 0.140 & 0.086 & 0.064 & 0.184 & 0.150 \\
\textsc{OpeNLGauge$_{ft}$} & \underline{0.416} & \textbf{0.498} & \textbf{0.464} & \textbf{0.391} & 0.319 & 0.460 & \textbf{0.425} \\
\bottomrule
\end{tabular}
\caption{Segment-level Spearman (\(\rho\)) correlations of different metrics on HANNA. \textbf{Coh.} = coherence, \textbf{Rel.} = relevance, \textbf{Eng.} = engagement, \textbf{Emp.} = empathy, \textbf{Sur.} = surprise, \textbf{Com.} = complexity, \textbf{Avg.} = average.}
\label{tab:meta_eval_hanna_rho}
\end{table*}

\begin{table*}
\centering \small
\begin{tabular}{lccccccc}
\toprule
\textbf{Metric} & \textbf{Coh.} & \textbf{Rel.} & \textbf{Eng.} & \textbf{Emp.} & \textbf{Sur.} & \textbf{Com.} & \textbf{Avg.} \\
\midrule
BLEU & 0.248 & 0.209 & 0.260 & 0.230 & 0.220 & 0.305 & 0.245 \\
ROUGE-1 & 0.287 & 0.237 & 0.306 & 0.260 & \underline{0.262} & 0.376 & 0.288 \\
METEOR & 0.278 & 0.224 & 0.303 & 0.269 & 0.261 & 0.377 & 0.285 \\
MoverScore & 0.289 & 0.280 & 0.308 & 0.242 & 0.236 & 0.353 & 0.285 \\
BERTScore & 0.273 & 0.257 & 0.304 & 0.260 & 0.234 & 0.348 & 0.279 \\
BARTScore & 0.185 & 0.177 & 0.209 & 0.206 & 0.164 & 0.212 & 0.192 \\
\midrule
\textsc{OpeNLGauge$_{ens}$} & 0.307 & 0.367 & 0.350 & 0.280 & 0.208 & 0.378 & 0.315 \\
 \textbullet \space Command R+ 104B & 0.270 & 0.297 & 0.300 & 0.253 & 0.171 & 0.277 & 0.261 \\
 \textbullet \space Gemma 2 27B & 0.329 & 0.328 & 0.343 & 0.267 & 0.241 & 0.384 & 0.315 \\
 \textbullet \space Llama 3.1 Nemotron 70B  & \textbf{0.369} & \underline{0.377} & \underline{0.364} & 0.272 & 0.221 & \underline{0.388} & \underline{0.332} \\
 \textbullet \space Mistral Large 2 123B  & 0.253 & 0.354 & 0.334 & \underline{0.301} & \textbf{0.284} & \textbf{0.405} & 0.322 \\
 \textbullet \space Qwen 2.5 72B & 0.294 & 0.360 & 0.307 & 0.222 & -0.003 & 0.353 & 0.255 \\
\midrule
Llama 3.1 8B & 0.079 & 0.285 & 0.116 & 0.073 & 0.055 & 0.152 & 0.127 \\
\textsc{OpeNLGauge$_{ft}$} & \underline{0.341} & \textbf{0.400} & \textbf{0.377} & \textbf{0.323} & 0.257 & 0.377 & \textbf{0.346} \\
\bottomrule
\end{tabular}
\caption{Segment-level Kendall (\(\tau\)) correlations of different metrics on HANNA. \textbf{Coh.} = coherence, \textbf{Rel.} = relevance, \textbf{Eng.} = engagement, \textbf{Emp.} = empathy, \textbf{Sur.} = surprise, \textbf{Com.} = complexity, \textbf{Avg.} = average.}
\label{tab:meta_eval_hanna_tau}
\end{table*}

\begin{table*}[t]
\centering \small
\begin{tabular}{lcccc}
\toprule
\textbf{Metric} & \textbf{Fluency} & \textbf{Meaning} & \textbf{Simplicity} & \textbf{Average} \\
\midrule
BLEU & 0.460 & 0.622 & 0.438 & 0.507 \\
SARI & 0.335 & 0.534 & 0.366 & 0.412 \\
BERTScore & 0.636 & 0.682 & 0.614 & 0.644 \\
\midrule
LENS & \underline{0.816} & 0.662 & 0.733 & 0.737 \\
\midrule
\textsc{OpeNLGauge$_{ens}$} & \textbf{0.840} & \textbf{0.864} & \textbf{0.770} & \textbf{0.825} \\
 \textbullet \space Command R+ 104B & 0.704 & 0.787 & 0.601 & 0.697 \\
 \textbullet \space Gemma 2 27B & 0.755 & 0.769 & 0.688 & 0.737 \\
 \textbullet \space Llama 3.1 Nemotron 70B & 0.778 & 0.822 & 0.660 & 0.753 \\
 \textbullet \space Mistral Large 2 123B & 0.705 & 0.744 & \underline{0.735} & 0.728 \\
 \textbullet \space Qwen 2.5 72B & 0.771 & 0.829 & 0.730 & 0.776 \\ 
\midrule
Llama 3.1 8B & 0.373 & 0.528 & 0.313 & 0.405 \\
\textsc{OpeNLGauge$_{ft}$} & 0.801 & \underline{0.851} & 0.716 & \underline{0.789} \\
\bottomrule
\end{tabular}
\caption{Segment-level Pearson (\(r\)) correlations of different metrics on Wiki-DA. \textbf{Meaning} = Meaning preservation.}
\label{tab:meta_eval_wikida}
\end{table*}

\begin{figure*}[h!]
    \centering
    \includegraphics[width=0.6\textwidth]{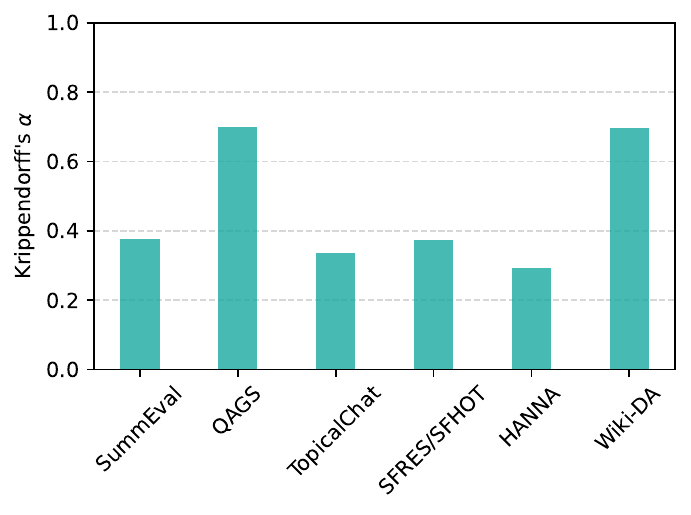}
    \caption{Inter-annotator agreement (Krippendorff's $\alpha$) between all LLMs in the ensemble for all meta-evaluation datasets. For each dataset, the coefficient is computed over all evaluation aspects.}
    \label{fig:ensemble_kripp}
\end{figure*}

\begin{table*}\small
\centering
\begin{tabular}{lcccccc}
\toprule
\textbf{Model/Method} & \textbf{QAGS} & \textbf{SummEval} & \textbf{TopicalChat} & \textbf{SFRES/SFHOT} & \textbf{HANNA} & \textbf{Wiki-DA} \\
\midrule
Command R+ 104B & 0.681 & 0.394 & 0.332 & 0.266 & 0.323 & 0.681 \\
Gemma 2 27B & 0.643 & 0.442 & 0.481 & 0.295 & 0.366 & 0.767 \\
Llama 3.1 Nemotron 70B & 0.669 & 0.451 & 0.660 & 0.179 & 0.393 & 0.752 \\
Mistral Large 2 123B & 0.645 & 0.502 & 0.598 & 0.249 & 0.376 & 0.769 \\
Qwen 2.5 72B & 0.651 & 0.478 & 0.496 & 0.283 & 0.301 & 0.783 \\
\midrule
Average & \textbf{0.688} & 0.533 & \textbf{0.652} & \textbf{0.299} & 0.409 & \textbf{0.837} \\
Average w/o outliers & 0.677 & \textbf{0.538} & 0.623 & 0.289 & \textbf{0.411} & 0.825 \\
Majority vote & 0.654 & 0.504 & 0.556 & 0.290 & 0.381 & 0.785 \\
Median & 0.668 & 0.509 & 0.594 & 0.296 & 0.401 & 0.803 \\
Min & 0.641 & 0.467 & 0.622 & 0.265 & 0.389 & 0.776 \\
\bottomrule
\end{tabular}
\caption{Segment-level Spearman (\(\rho\)) correlations of different score aggregation methods. For each dataset, correlations are averaged across aspects. Individual LLMs are included for comparison. Best \emph{ensemble} results for each dataset are highlighted in bold.}
\label{tab:aggregation_methods}
\end{table*}

\begin{figure*}[h]
    \centering\small
    \begin{tikzpicture}[align=left]

        \node[draw=gray!30, rectangle, line width=0.5mm, text width=15.5cm, anchor=north west, align=left, fill=gray!7.5, rounded corners] (context) at (-7.5, 4) {
            \textbf{Data:}\\
            \texttt{Adolfo\_Suárez\_Madrid–Barajas\_Airport | operatingOrganisation | ENAIRE \\
Adolfo\_Suárez\_Madrid–Barajas\_Airport | runwayLength | 3500.0 \\
Adolfo\_Suárez\_Madrid–Barajas\_Airport | elevationAboveTheSeaLevel | 610.0 \\
Adolfo\_Suárez\_Madrid–Barajas\_Airport | runwayName | "14L/32R"} \\[0.1cm]
            \textbf{Text:} Adolfo Suárez Madrid–Barajas Airport is a \hlc[error2]{major international airport located near the heart of the city}\textsuperscript{[1]}, \\\hlc[error1]{offering a wide range of air services}\textsuperscript{[2]}.
        };

        \node[anchor=north west] (boxcontainer) at ([yshift=-0.3cm, xshift=0.1cm]context.south west) {
                    \textbf{Evaluation aspect:} \textit{Faithfulness} -- Extent to which the information in the text is supported by the
                    data
        }; 

        \node[draw=error2_border, rectangle, line width=0.25mm, text width=7.3cm, below=0.6cm of boxcontainer.north west, anchor=north west, align=left, fill=error2, xshift=3pt] (box1) {
            \textbf{[1] Explanation:} This phrase is considered an error because the provided data does not mention the airport's size (major/international), location relative to the city (near the heart), or any information that could imply these details. The text introduces unsupported information, negatively affecting faithfulness. \\
            \textbf{Severity:} 4
        };

        \node[draw=error1_border, rectangle, line width=0.25mm, text width=7.3cm, right=0.2cm of box1.north east, anchor=north west, align=left, fill=error1] (box2) {
            \textbf{[2] Explanation:} Similar to Error 1, this phrase introduces information not present in the provided data. There is no mention of the types or variety of air services offered by the airport, making this an unsupported claim that affects faithfulness. \\
            \textbf{Severity:} 3
        };

        \node[rectangle, text width=15cm, below=0.2cm of box1.south west, anchor=north west, align=left] (score) {
            \textbf{Overall score:} Poor (2/5) \\[0.2cm]
            \textbf{Explanation of the score:} The generated text contains significant errors related to faithfulness, as it introduces substantial information not supported by the provided data. While the text does not contradict any data points (since it doesn't directly address them), its lack of adherence to the given facts and introduction of unsubstantiated claims warrant a `Poor' rating. For an `Excellent' score, the text should have strictly adhered to the provided data, which includes specific details about the airport's operating organisation, runway length, elevation, and runway name, none of which were accurately represented or utilized in the generated text.
        };

        \draw[thick, color=gray!60] ([xshift=-5pt, yshift=0.7cm] box1.north west) 
            rectangle 
            ([xshift=8pt, yshift=-5pt] score.south east);
        
    \end{tikzpicture}
    \caption{Example error span annotation provided by \textsc{OpeNLGauge} for the RDF-to-text task. The generated text is evaluated for \emph{faithfulness} with respect to the input data.}
    \label{fig:exp2}
\end{figure*}

\begin{figure*}[h]
    \centering\small
    \begin{tikzpicture}[align=left]

        \node[draw=gray!30, rectangle, line width=0.5mm, text width=15.5cm, anchor=north west, align=left, fill=gray!7.5, rounded corners] (context) at (-7.5, 4) {
            \textbf{Emotional Context:} A is guilty\\
            \textbf{Dialogue History:}\\
            A: One time I skipped my friend's birthday party to watch TV\\
            B: I hope something good was on\\
            A: There was! But I felt so bad afterwards. I'm a monster\\[0.1cm]
            \textbf{Response:} \hlc[error1]{It's okay}\textsuperscript{[1]}, everyone makes mistakes sometimes. \hlc[error2]{Next time, try to find a balance between social commitments and personal enjoyment}\textsuperscript{[2]}.!
        };

        \node[anchor=north west] (boxcontainer) at ([yshift=-0.4cm, xshift=0.1cm]context.south west) {
            \textbf{Evaluation aspect:} \textit{Empathy} -- Extent to which the response shows understanding of the feelings of~the person talking\\ about their experience
        }; 

        \node[draw=error1_border, rectangle, text width=7.3cm, below=1.0cm of boxcontainer.north west, anchor=north west, align=left, line width=0.25mm, fill=error1, xshift=3pt] (box1) {
            \textbf{[1] Explanation:} While the intention is to comfort, the phrase "It's okay" might come across as dismissive of A's expressed guilt and remorse, potentially undermining their emotional experience. A more empathetic approach could acknowledge the negative feelings before offering reassurance.\\
            \textbf{Severity:} 2
        };

        \node[draw=error2_border, rectangle, text width=7.3cm, right=0.2cm of box1.north east, anchor=north west, align=left, line width=0.25mm, fill=error2] (box2) {
            \textbf{[2] Explanation:} This part of the response shifts focus from acknowledging and validating A's current emotional state (guilt) to providing advice for future actions. While helpful, it doesn't directly address or show understanding of A's feelings about their past action, slightly lacking in empathy.\\
            \textbf{Severity:} 3
        };

        \node[rectangle, text width=15cm, below=0.2cm of box1.south west, anchor=north west, align=left] (score) {
            \textbf{Overall score:} Fair (3/5) \\[0.2cm]
            \textbf{Explanation of the score:} The response demonstrates some level of empathy by attempting to reassure and offer a solution, indicating an understanding that mistakes happen. However, it falls short in fully acknowledging and validating the speaker's emotional experience (guilt), hence not scoring higher than 'Fair'. To achieve a 'Good' or 'Excellent' score, the response would need to more directly address and empathize with A's feelings of guilt and self-perceived monstrosity.
        };

        \draw[thick, color=gray!60] ([xshift=-5pt, yshift=1.2cm] box1.north west) 
            rectangle 
            ([xshift=9pt, yshift=-5pt] score.south east);
        
    \end{tikzpicture}
    \caption{Example error span annotation provided by \textsc{OpeNLGauge} for the dialogue response generation task. The response is evaluated for \emph{empathy}. In addition to dialogue history, the input also includes the emotional context of the person talking about their experience.}
    \label{fig:exp3}
\end{figure*}

\begin{figure*}[h]
    \centering\small
    \begin{tikzpicture}[align=left]

        \node[draw=gray!30, rectangle, line width=0.5mm, text width=15.5cm, anchor=north west, align=left, fill=gray!10, rounded corners] (context) at (-7.5, 4) {
            \textbf{Source:} \emph{[NOT PROVIDED]}\\
            \textbf{Summary:} \hlc[error2]{By A new study that college are more likely to engage in energy - binge drinking when alcohol than non - .}\textsuperscript{[1]}\\
        };

        \node[anchor=north west] (boxcontainer) at ([yshift=-0.4cm, xshift=0.1cm]context.south west) {
            \textbf{Evaluation aspect:} \textit{Fluency} -- Formal quality of individual sentences of the summary. A fluent sentence should be\\ grammatical, natural and easy to understand.
        }; 

        \node[draw=error2_border, line width=0.25mm, rectangle, text width=15cm, below=1.0cm of boxcontainer.north west, anchor=north west, align=left, fill=error2, xshift=3pt] (box1) {
            \textbf{[1] Explanation:} The sentence is fragmented and lacks proper structure. It starts with an unnecessary preposition ("By") and contains several grammatical errors, such as missing words and incorrect phrasing. This makes the sentence difficult to understand and unnatural.\\
            \textbf{Severity:} 5
        };

        \node[rectangle, text width=15cm, below=0.2cm of box1.south west, anchor=north west, align=left] (score) {
            \textbf{Overall score:} Unacceptable (1/5) \\[0.2cm]
            \textbf{Explanation of the score:} The summary is very disfluent due to significant grammatical errors and lack of coherence, making it extremely difficult to comprehend.
        };

        \draw[thick, color=gray!60] ([xshift=-5pt, yshift=1.2cm] box1.north west) 
            rectangle 
            ([xshift=9pt, yshift=-5pt] score.south east);
        
    \end{tikzpicture}
    \caption{Example error span annotation provided by \textsc{OpeNLGauge} for the summarization task. The summary is evaluated for \emph{fluency}. Note that source text is not included, as it is not relevant for evaluation of fluency.}
    \label{fig:exp4}
\end{figure*}

\end{document}